\documentclass[journal]{IEEEtran}

\usepackage[noadjust]{cite} % 添加 noadjust 避免多余空格
\usepackage{graphicx}
\usepackage{caption}
\usepackage{subcaption}
\usepackage{booktabs}
\usepackage{multirow}
\usepackage{adjustbox}
\usepackage{amsmath, amssymb}
\usepackage{array}
\usepackage{enumitem}
\usepackage{algorithm}
\usepackage{algorithmic}
\usepackage{caption}
\begin{document}
\title{Multi-Task Genetic Algorithm with Multi-Granularity Encoding for Protein-Nucleotide Binding Site Prediction}

\author{Yiming Gao, Liuyi Xu, Pengshan Cui, Yining Qian*, An-Yang Lu, Xianpeng Wang%
\thanks{This work was supported in part by the National Natural Science Foundation of China (NSFC) under Grant Nos. 62522309, and in part by the Natural Science Foundation of Liaoning Province under Grant No. 2025JH6/101000012. (Corresponding author: Yining Qian.)}%
\thanks{This work has been submitted to the IEEE for possible publication. Copyright may be transferred without notice, after which this version may no longer be accessible.}%
\thanks{Yiming Gao, Liuyi Xu, Pengshan Cui and An-Yang Lu are with the College of Information Science and Engineering, Northeastern University, Shenyang, 110819, China (e-mail: gaoym3@mails.neu.edu.cn; xuliuyi@mails.neu.edu.cn; ccuipengshan@163.com; luanyang@mail.neu.edu.cn).}%
\thanks{Yining Qian is with the School of Computer Science and Engineering, Northeastern University, Shenyang 110819, China (e-mail: qianyiningning@126.com). Corresponding author.}%
\thanks{Xianpeng Wang is with the Key Laboratory of Data Analytics and Optimization for Smart Industry, Ministry of Education, Northeastern University, Shenyang, 110819, China (wangxianpeng@ise.neu.edu.cn).}%
}
\maketitle

\begin{abstract}
Accurate identification of protein-nucleotide binding sites is fundamental to deciphering molecular mechanisms and accelerating drug discovery. However, current computational methods often struggle with suboptimal performance due to inadequate feature representation and rigid fusion mechanisms, which hinder the effective exploitation of cross-task information synergy. To bridge this gap, we propose MTGA-MGE, a framework that integrates a Multi-Task Genetic Algorithm with Multi-Granularity Encoding to enhance binding site prediction. Specifically, we develop a Multi-Granularity Encoding (MGE) network that synergizes multi-scale convolutions and self-attention mechanisms to distill discriminative signals from high-dimensional, redundant biological data. To overcome the constraints of static fusion, a genetic algorithm is employed to adaptively evolve task-specific fusion strategies, thereby effectively improving model generalization. Furthermore, to catalyze collaborative learning, we introduce an External-Neighborhood Mechanism (ENM) that leverages biological similarities to facilitate targeted information exchange across tasks. Extensive evaluations on fifteen nucleotide datasets demonstrate that MTGA-MGE not only establishes a new state-of-the-art in data-abundant, high-resource scenarios but also maintains a robust competitive edge in rare, low-resource regimes, presenting a highly adaptive scheme for decoding complex protein-ligand interactions in the post-genomic era.
\end{abstract}


\section{Introduction}
\begin{figure}[!htbp]
    \centering
    \includegraphics[width=0.8\linewidth]{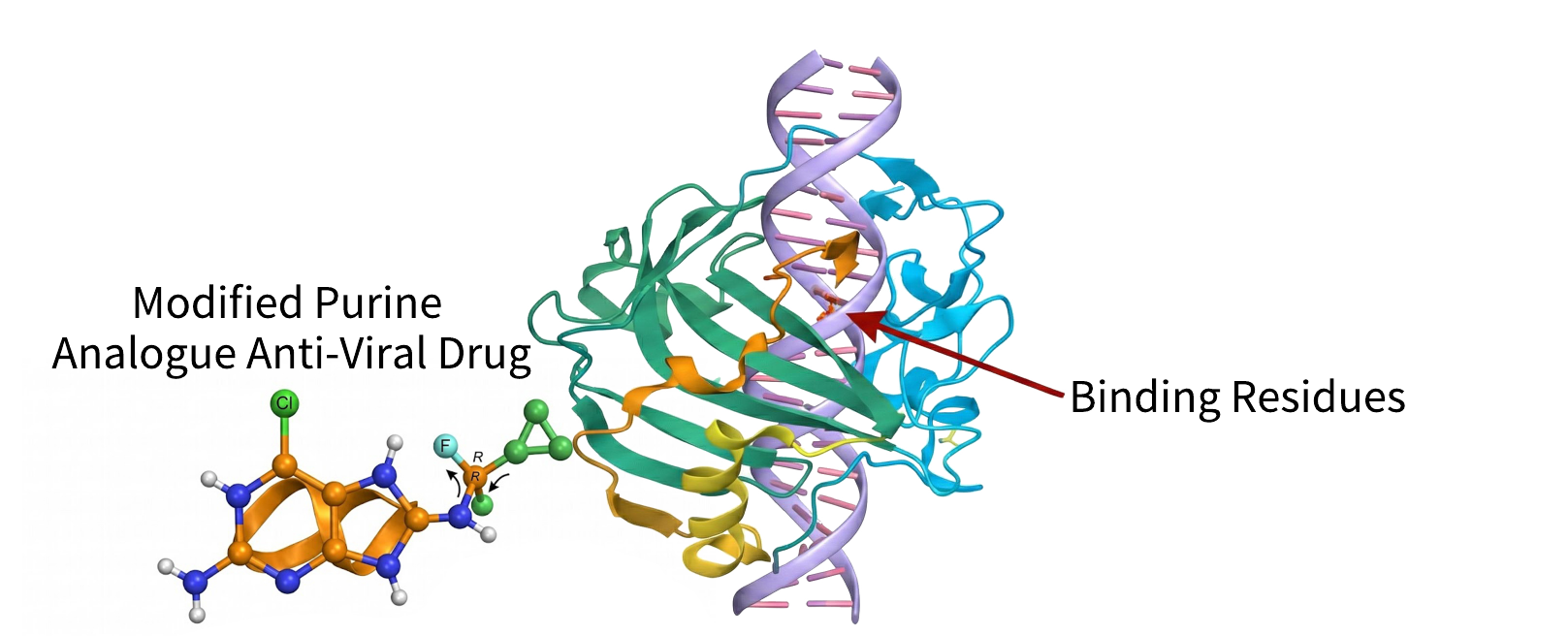}
    \caption{Schematic representation of protein–nucleotide interaction, further illustrating the action of a drug at the protein–nucleotide binding sites. The highlighted regions (orange/yellow) illustrate the binding residues distributed across the protein structure.}
    \label{fig:binding_challenge}
\end{figure}

Protein-nucleotide binding site prediction is a cornerstone of molecular biology, aiming to pinpoint specific residues that interact with ligands such as DNA, RNA, and various nucleotides, thereby elucidating essential biological processes and guiding structural drug design~\cite{Chandel2021Nucleotide,Yan2016DNARNABinding,Alemasova2017RNA,Berdis2022Nucleobase}. As visualized in Fig. \ref{fig:binding_challenge}, 
the spatial arrangement of these residues defines the interaction interface, imposing strict constraints for protein function annotation and nucleotide-targeted drug discovery~\cite{Blass2015DrugDiscovery,Valkov2011Fragment,rhodes2024targeting}. Without high-precision site identification, elucidating complex biological mechanisms at the molecular level remains elusive, while the lack of functional annotation hinders the efficiency of drug discovery~\cite{vanTienhoven2025DarkProteome}.

Although experimental techniques such as X-ray crystallography, site-directed mutagenesis, and cross-linking provide reliable evidence~\cite{Sagendorf2017DNAproDB,Boutureira2015ChemicalModification,Stutzer2020Crosslinking}, their high costs and low throughput render them inadequate for the exponentially growing volume of protein sequences. Consequently, data-driven computational approaches have emerged as the dominant research paradigm~\cite{10.1093/g3journal/jkaf169}.
    
Current computational methods can be broadly categorized into nucleotide-specific models~\cite{Chauhan2010GTPbinder,Le2019DeepGTP,Hu2021DeepATPseq} and nucleotide-general models~\cite{Ding2017ECRUS,Zhu2019BGSVM,Zhao2019SXGBsite,Ding2022GHKNN,Wu2025NucGMTL,wu2025identification}. Nucleotide-specific models, such as DeepGTP~\cite{Le2019DeepGTP} and DeepATPseq~\cite{Hu2021DeepATPseq}, typically employ elaborate feature engineering for a specific ligand. While achieving acceptable accuracy on specific tasks, their critical limitation lies in their ``isolated modeling'' paradigm. By treating each nucleotide-binding event as an independent problem, these methods completely fail to leverage the shared biological homologies across different ligands~\cite{wu2025identification}. Ignoring these cross-task homologies leads to an excessive reliance on task-specific data, restricting generalizability and resulting in suboptimal performance in low-resource scenarios where labeled samples for niche ligands are scarce~\cite{pourmirzaei2025zero}.

To overcome this limitation, recent research has pivoted toward nucleotide-general models utilizing Multi-Task Learning (MTL). For instance, approaches like NucGMTL~\cite{Wu2025NucGMTL} and NucMoMTL~\cite{wu2025identification} employ multi-task learning to jointly train on diverse nucleotide-binding tasks, leveraging shared biological rules to benefit low-resource ligands~\cite{wang2025mpbind}. Furthermore, these methods increasingly rely on Pre-trained Protein Language Models (PLMs)~\cite{Brandes2022ProteinBERT,Elnaggar2021ProtTrans,Lin2023ESM2} for sequence feature extraction. While PLMs provide rich global context, their extremely high-dimensional embeddings~\cite{van2026plm} are often processed using simple linear methods (e.g., mean pooling). This ``coarse-grained'' extraction obscures critical binding features with high-dimensional redundant noise~\cite{naderializadeh2025aggregating}, reducing sensitivity to weak local signals. Similarly, the simplistic concatenation of traditional handcrafted features also exacerbates dimensional redundancy and feature conflicts, further drowning out these weak signals. Because nucleotide binding is determined by subtle spatial arrangements of local residues, losing these faint signals prevents models from precisely pinpointing binding anchors within complex backgrounds~\cite{wang2025mpbind}.

Beyond feature-level representation, the structural integration of these features presents a further challenge. Existing nucleotide-general models predominantly adopt a shared-backbone fusion paradigm~\cite{pourmirzaei2025prot2token}. Although parameter sharing formally induces inter-task correlation, fixed, manually designed architectures fail to dynamically accommodate the biological heterogeneity across diverse nucleotide tasks~\cite{liu2019loss}. Consequently, models are forced to compromise conflicting ligand demands through constrained weight fine-tuning. Mathematically, discovering task-specific optimal fusion strategies is a non-differentiable combinatorial optimization problem~\cite{Liu2021ENAS_Survey} where traditional gradient-based algorithms fail~\cite{xue2025dominant}. Thus, the crux lies in the structural inability of fixed architectures to autonomously optimize fusion strategies.

Addressing this bottleneck necessitates shifting from manually designed architectures~\cite{Wu2025NucGMTL} to automated, task-specific architecture search~\cite{danneels2025mtl}. While Evolutionary Algorithms (EAs) excel at such non-differentiable optimization problems~\cite{Liu2021ENAS_Survey}, their deployment within MTL presents critical challenges. Existing MTL approaches often suffer from restricted shallow sharing, which lacks explicit channels for knowledge transfer~\cite{Zhang2022PanSpecific,Li2022DeepMapi,Zhang2023CrossTask,Chen2024GraphMultiTask}. This hinders the cross-task exchange of advantageous architectural motifs among homologous targets, making it profoundly difficult to orchestrate complex interplays and avoid negative interference across diverse tasks. Therefore, constructing an efficient mechanism that simultaneously searches for optimal architectures and explicitly exploits biological homologies for cross-task ``co-evolution'' remains a crucial, unresolved hurdle for maximizing predictive robustness~\cite{zhang2024constrained}.

To address these challenges, this study develops the Multi-Task Genetic Algorithm with Multi-Granularity Encoding (MTGA-MGE), a unified framework specifically engineered to enhance protein-nucleotide binding site prediction. By synergizing adaptive architecture evolution with refined feature representation, MTGA-MGE systematically resolves the bottlenecks of rigid fusion and information redundancy. The main contributions of this work are as follows:

(i) We design a Multi-Granularity Encoding (MGE) network to mitigate noise and redundancy inherent in high-dimensional embeddings. By synergizing multi-scale convolutions with self-attention mechanisms, MGE effectively distills discriminative local signals and captures global dependencies, thereby providing a highly refined representation for deciphering complex biological interactions.

(ii) We propose an NSGA-III-based Multi-Task Genetic Algorithm (MTGA) to bypass non-differentiable optimization constraints and overcome the limitations of static designs. This approach enables the adaptive evolution of task-specific feature fusion strategies, which improves generalization across diverse and complex binding scenarios.

(iii) We introduce an External-Neighborhood Mechanism (ENM) to catalyze targeted information exchange and facilitate efficient cross-task collaboration. By quantifying physicochemical homology to bridge task-specific channels, ENM strengthens co-evolution between tasks, ensuring robust predictive performance across diverse binding scenarios.

\section{Related Work}

\subsection{Computational Methods for Protein-Nucleotide Binding Site Prediction}

The evolution of protein-nucleotide binding site prediction has transitioned from feature-engineered machine learning to end-to-end deep learning paradigms. Pioneering efforts focused on manual feature construction. For instance, EC-RUS~\cite{Ding2017ECRUS} integrated evolutionary profiles with physicochemical properties via Random Forest classifiers, effectively enhancing representation diversity beyond simple sequence information. To ameliorate the pervasive class imbalance, BGSVM-NUC~\cite{Zhu2019BGSVM} employed Granular Support Vector Machines coupled with granular under-sampling techniques, effectively reducing model bias toward non-binding residues. With the advent of deep learning, research expanded into nucleotide-specific and nucleotide-general models. Nucleotide-specific models, such as DeepGTP~\cite{Le2019DeepGTP} and DeepATPseq~\cite{Hu2021DeepATPseq}, leverage Convolutional Neural Networks (CNNs) to automatically extract deep local features, achieving superior precision for designated GTP and ATP targets, respectively. To broaden applicability, nucleotide-general models have emerged. SXGBsite~\cite{Zhao2019SXGBsite} combines Extreme Gradient Boosting with SMOTE over-sampling to consolidate PSSM and solvent accessibility features, ensuring efficient and robust prediction. GHKNN~\cite{Ding2022GHKNN} introduces a Graph Hashing-based K-Nearest Neighbors algorithm, which explicitly encodes topological dependencies between residues via graph structures to enhance spatial representation. NucGMTL~\cite{Wu2025NucGMTL} utilizes PLMs within a Grouped MTL framework, successfully extracting shared commonalities across diverse nucleotides to improve generalization.

\begin{figure*}[htbp]
    \centering
    \includegraphics[width=\textwidth, trim=4cm 4cm 5cm 0.3cm, clip]{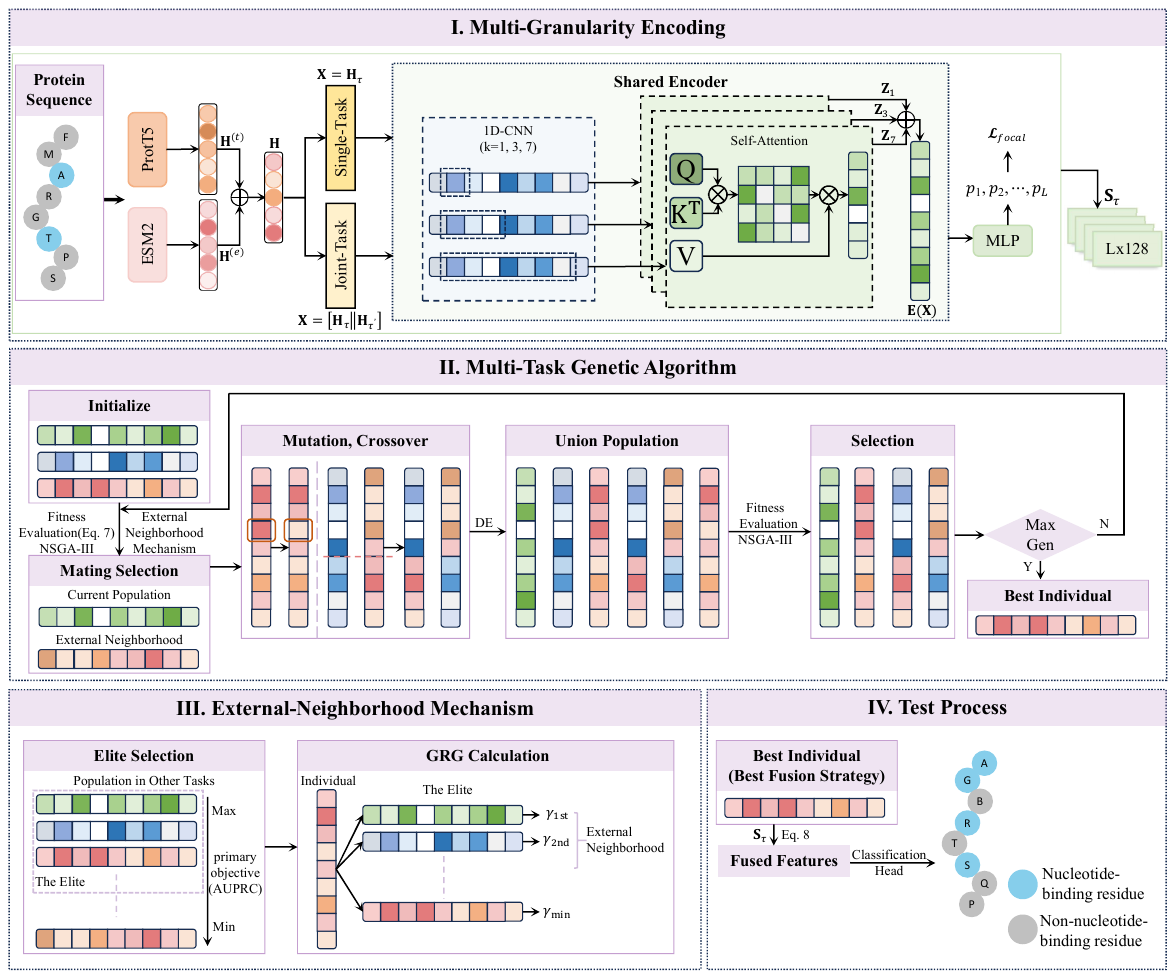}
    \caption{Framework of MTGA-MGE. The framework operates through three key components: I. Multi-Granularity Encoding, II. Multi-Task Genetic Algorithm, III. External-Neighborhood Mechanism. Finally, IV. Test Process illustrates the model inference phase.}
    \label{fig:MTGA-MGE-framework}
    \rule{\textwidth}{0.75pt}
\end{figure*}

Nucleotide-specific models suffer from task isolation, precluding cross-task knowledge transfer and hindering performance on data-scarce ligands~\cite{liu2026dtap}. Conversely, although nucleotide-general models incorporate MTL, they predominantly rely on fixed, manually designed architectures. This static paradigm fails to efficiently navigate the vast discrete combinatorial space to identify optimal, task-specific fusion strategies~\cite{long2025enhancing}. These limitations underscore the urgent need to shift from manual architectural engineering to automated architecture search, which motivates our work.

\subsection{Evolutionary Algorithms}

EAs, particularly Genetic Algorithms (GAs), have demonstrated robust global optimization capabilities across diverse complex systems, maintaining their vanguard status in modern deep learning optimization. Recently, the efficacy of EAs within bioinformatics has gained significant traction. For instance, Qian et al.~\cite{Qian2025ProteinSS} utilized evolutionary strategies to optimize deep fusion architectures for protein secondary structure prediction, demonstrating the superiority of automated architecture search over manual design. However, extrapolating these methodologies to protein-nucleotide binding site prediction presents unique challenges. Unlike the single-task nature of protein secondary structure prediction, nucleotide binding prediction necessitates MTL across diverse ligand types, thereby inducing an exponential expansion of the architecture search space~\cite{li2026contribution}. To navigate such high-dimensional and discrete combinatorial landscapes, NSGA-III~\cite{Jiang2023MOEAS,Zhu2024Evolutionary,Zhao2025AdaptiveNSGAIII,shaokuan} employs a reference-point-based selection mechanism. This mechanism ensures a uniform distribution of solutions along the Pareto front regardless of the search space's continuity. Consequently, this mathematical property renders NSGA-III highly amenable to the discrete characteristics of the feature fusion problem addressed in this study~\cite{Zhang2025NSGA3Discrete}.

\section{Methods}

This section provides an overview of the MTGA-MGE framework, followed by a detailed explanation of its three key components: the Multi-Granularity Encoding network, the Multi-Task Genetic Algorithm, and the External-Neighborhood Mechanism. The section concludes with the description of the test process for final inference.

\subsection{Framework of MTGA-MGE}

The MTGA-MGE framework (Fig.~\ref{fig:MTGA-MGE-framework}) comprises three core components: Multi-Granularity Encoding (MGE), Multi-Task Genetic Algorithm (MTGA), and External-Neighborhood Mechanism (ENM).

The MGE module constructs robust feature representations to overcome the challenges of high dimensionality and noise in features. It integrates multi-scale convolution with scale-based self-attention and employs a multi-task context injection mechanism. By doing so, it effectively distills discriminative residue-level patterns from redundant embeddings into a comprehensive candidate pool, enabling the model to more precisely pinpoint binding anchors within complex backgrounds.

The MTGA module automates the architecture search process by formulating feature fusion as a combinatorial optimization problem. Utilizing the NSGA-III framework, it executes a series of evolutionary operations—such as crossover and mutation—to iteratively optimize the population of fusion strategies. Consequently, it adaptively navigates a complex search space to identify task-specific configurations defined by feature subsets, operators, and weights. This approach successfully addresses the inherent rigidity of manual designs, thereby effectively improving the overall generalization ability of the model.

The ENM module introduces a collaborative knowledge interaction mechanism to enhance search efficiency and population diversity. It quantifies structural similarity via Gray Relational Analysis (GRA) to construct a dynamic reservoir of elite architectures from related tasks. Consequently, it explicitly guides the mutual exchange of valid structural blueprints, enabling efficient cross-task co-evolution.

\subsection{MGE Method}
\label{subsec:encoder}

The MGE method comprises four main components: dual-view protein representation, multi-granularity feature extraction, loss function for class imbalance, and multi-task context injection.

\textit{1) Dual-view Protein Representation}

To construct a comprehensive feature space, we first employ ProtT5~\cite{Elnaggar2021ProtTrans} to capture global contextual information. This yields a robust semantic embedding matrix denoted as $\mathbf{H}^{(t)}\in\mathbb{R}^{L\times d_t}$ ($d_t=1024$). Complementarily, we utilize ESM2~\cite{Lin2023ESM2} to extract features rich in evolutionary signals. This model generates a high-dimensional structural representation denoted as $\mathbf{H}^{(e)}\in\mathbb{R}^{L\times d_e}$ ($d_e=1280$).

To synthesize these complementary perspectives, we concatenate the two embeddings along the channel dimension to form a unified dual-view representation $\mathbf{H} \in \mathbb{R}^{L \times d_{in}}$, where $d_{in} = d_t + d_e =2304$. This integration effectively synergizes the global semantic information from ProtT5 with the evolutionary structural patterns from ESM2, establishing a solid foundation for the subsequent MGE.

\textit{2) Multi-Granularity Feature Extraction}

To construct a diverse candidate pool and distill task-relevant patterns, we first define a dual-task context injection strategy to form the input tensor. Let $\mathcal{T}$ denote the set of 15 nucleotide-binding tasks, specifically $\mathcal{T} = \{\text{ADP, AMP, ATP, \dots, UTP}\}$. Let $\mathbf{H}_{\tau} \in \mathbb{R}^{L \times d_{in}}$ ($\tau \in \mathcal{T}$) and $\mathbf{H}_{\tau'} \in \mathbb{R}^{L' \times d_{in}}$ ($\tau' \in \mathcal{T}$ (where $\tau\neq\tau'$)) denote the unified dual-view feature matrices for the primary task $\tau$ and an auxiliary task $\tau'$, respectively. We distinguish between single-task inputs ($\mathbf{X} = \mathbf{H}_{\tau} \in \mathbb{R}^{L \times d_{in}}$) and dual-task inputs constructed by concatenating feature sets along the sequence dimension: $\mathbf{X} = [\mathbf{H}_{\tau} \parallel \mathbf{H}_{\tau'}] \in \mathbb{R}^{(L+L') \times d_{in}}$.

This high-dimensional input tensor $\mathbf{X}$ is then processed through the MGE module to extract multi-granularity features. The tensor is routed through three parallel branches, each characterized by a distinct kernel size $k \in \{1, 3, 7\}$ to capture local patterns at varying receptive fields. Unlike standard architectures that fuse multi-scale features prior to context modeling, our design integrates a 1D convolutional layer ($d_{cnn} = 128$) followed directly by a specific Multi-Head Self-Attention (MHSA) module within each branch. Here, the MHSA employs the standard Transformer attention mechanism to capture long-range contextual dependencies across the sequence based on the extracted local representations. Reinforced by residual connections, the scale-specific refined features $\mathbf{Z}_k \in \mathbb{R}^{L \times 128}$ for each branch are computed as follows:
\begin{equation}
\mathbf{Z}_k = \mathrm{MHSA}(\mathrm{Conv1D}_k(\mathbf{X})) + \mathrm{Conv1D}_k(\mathbf{X})
\end{equation}
where $k \in \{1, 3, 7\}$. This structure allows the model to capture long-range dependencies specifically tailored to different local granularities. Finally, the refined representations from all three branches are concatenated along the channel dimension and fused via a linear projection to produce the final deep representation $\mathbf{E}(\mathbf{X}) \in \mathbb{R}^{L \times 128}$:
\begin{equation}
\mathbf{E}(\mathbf{X}) = \mathrm{Linear}(\mathbf{Z}_1 \oplus \mathbf{Z}_3 \oplus \mathbf{Z}_7)
\end{equation}
where $\oplus$ denotes the concatenation operation. This compact output serves as the robust feature basis for the subsequent evolutionary optimization stage.

\textit{3) Loss Function for Class Imbalance}

Given the extreme class imbalance between binding and non-binding residues, standard cross-entropy loss tends to bias the model toward the majority negative class. We therefore adopt the Focal Loss~\cite{Lin2017FocalLoss} to dynamically down-weight easy negatives and focus training on hard examples. For a residue with predicted probability $p\in[0,1]$ and ground truth label $y\in\{0,1\}$, the loss is defined as:
\begin{equation}
\mathcal{L}_{\text{focal}} = -\alpha\, y(1-p)^{\gamma}\log(p) - (1-\alpha)(1-y)p^{\gamma}\log(1-p)
\end{equation}
where hyperparameters $\alpha$ and $\gamma$ dynamically down-weight easy negatives and focus training on hard examples, effectively improving sensitivity for rare binding residues.

\textit{4) Dual-task Context Injection}

Single-task learning often ignores shared structural and evolutionary patterns across different nucleotide-binding tasks. To address this, we introduce a dual-task context injection strategy. By incorporating an auxiliary task, the encoder synthesizes complementary information, allowing the deep representation $\mathbf{E}(\mathbf{X})$ (defined in Eq. 2) to implicitly capture broader structural motifs. This constructs a diverse and information-rich candidate pool for the subsequent evolutionary search.

\begin{algorithm}[h]
\caption{Framework of MTGA}
\label{alg:main_framework}
\renewcommand{\algorithmicrequire}{\textbf{Input:}}
\renewcommand{\algorithmicensure}{\textbf{Output:}}
\begin{algorithmic}[1]
\REQUIRE 
    Data $\mathcal{D}_{tr}, \mathcal{D}_{val}$; Population size $N$; Max generations $G_{max}$.
\ENSURE Pareto optimal set for each task.
\FOR{each task $t \in \mathcal{T}$}
    \STATE Initialize $P_t$ and evaluate fitness $F(\mathbf{x})$;
\ENDFOR
\STATE Initialize NSGA-III reference points $Z$;
\FOR{$g = 1$ to $G_{max}$}
    \STATE $\mathbf{N} \leftarrow$ Construct External-Neighborhoods via \textbf{Algorithm~\ref{alg:neighborhood}};
    \FOR{each task $t \in \mathcal{T}$}
        \STATE Retrieve $\mathcal{N}_t$ from $\mathbf{N}$;
        \STATE $Q_t \leftarrow \emptyset$;
        \WHILE{$|Q_t| < N$}
            \STATE $c \leftarrow$ Generate Offspring via \textbf{Algorithm~\ref{alg:variation}};
            \STATE $Q_t \leftarrow Q_t \cup \{c\}$;
        \ENDWHILE
        \STATE Apply Batch DE on weights of $Q_t$; 
        \STATE Evaluate fitness $F(Q_t)$;
        \STATE $R_t \leftarrow P_t \cup Q_t$;
        \STATE Perform Non-dominated Sort on $R_t$ to get fronts;
        \STATE Normalize objectives and associate with $Z$;
        \STATE $P_t \leftarrow$ Select next generation from $R_t$ via Niche-Preservation;
    \ENDFOR
\ENDFOR
\RETURN Non-dominated individuals from final populations;
\end{algorithmic}
\end{algorithm}

For a target task $\tau \in \mathcal{T}$, we generate a comprehensive feature pool containing one single-task representation and 28 dual-task representations by pairing $\tau$ with every other nucleotide task $\tau' \in \mathcal{T}$ (where $\tau\neq\tau'$). We explicitly employ a bidirectional pairing scheme—constructing both $(\tau, \tau')$ and $(\tau', \tau)$—to capture asymmetric inter-task correlations, in each pair, only the first element (the primary task) serves as the supervision target to compute the training loss, whereas the second element (the auxiliary task) merely provides contextual features and does not participate in the loss calculation. Formally, the candidate input set $\mathcal{X}_{pool}$ is defined as:
\begin{equation}
\mathcal{X}_{pool} = \{ \mathbf{H}_{\tau} \} \cup \bigcup_{\tau' \in \mathcal{T} \setminus \{\tau\}} \left\{ [\mathbf{H}_{\tau} \parallel \mathbf{H}_{\tau'}], [\mathbf{H}_{\tau'} \parallel \mathbf{H}_{\tau}] \right\}
\end{equation}
The resulting set of 29 candidate deep representations is thus given by:
\begin{equation}
\mathcal{S}_{\tau} = \{ \mathbf{E}(\mathbf{X}) \mid \mathbf{X} \in \mathcal{X}_{pool} \}
\end{equation}

\subsection{MTGA Method}
\label{subsec:optimization}
To identify optimal task-specific feature fusion strategies, we formulate the learning problem for each nucleotide task as a bi-objective optimization. This problem inherently represents a mixed-integer programming challenge, involving both discrete combinatorial optimization (feature subset selection and operator arrangement) and continuous parameter optimization (fusion weight fine-tuning).

\textit{1) Bi-objective Formulation}

\begin{algorithm}[h]
\caption{Offspring generation strategy}
\label{alg:variation}
\renewcommand{\algorithmicrequire}{\textbf{Input:}}
\renewcommand{\algorithmicensure}{\textbf{Output:}}
\begin{algorithmic}[1]
\REQUIRE 
    Current population $P_t$; External-Neighborhood $\mathcal{N}_t$;
    Transfer prob. $P_{ext}$; Crossover rate $P_c$; Mutation rate $P_m$;
    Specific probs $\{P_{struct}, P_{op}, P_{weight}\}$.
\ENSURE A generated offspring $c$.
\STATE Select first parent $p_1$ from $P_t$ via Tournament Selection;
\IF{$rand() < P_{ext}$ \AND $\mathcal{N}_{p_1} \neq \emptyset$}
    \STATE Select second parent $p_2$ from $\mathcal{N}_{p_1}$;
\ELSE
    \STATE Select second parent $p_2$ from $P_t$;
\ENDIF
\STATE $c \leftarrow p_1$.copy();
\IF{$rand() < P_c$} 
    \STATE $c \leftarrow$ SinglePointCrossover$(p_1, p_2)$;
\ENDIF
\IF{$rand() < P_m$}
    \IF{$rand() < P_{struct}$}
        \STATE Apply Structural Mutation;
    \ENDIF
    \IF{$rand() < P_{op}$}
        \STATE Apply Operator Mutation;
    \ENDIF
    \IF{$rand() < P_{weight}$}
        \STATE Apply Weight Mutation (using $\mathcal{N}_{p_1}$);
    \ENDIF
\ENDIF
\RETURN $c$;
\end{algorithmic}
\end{algorithm}

To efficiently evaluate fusion strategies, we employ a proxy mechanism where a logistic regression head is trained on standardized features for each individual $\mathbf{x}$. This classifier utilizes the Focal Loss (defined in Eq. 3) to mitigate class imbalance. We formulate the optimization as maximizing the validation Area Under the Precision-Recall Curve (AUPRC, defined in Eq. 11) while minimizing the False Positive Rate (FPR). The FPR is defined as:
\begin{equation}
\text{FPR} = \frac{FP}{FP + TN}
\end{equation}
where $FP$ and $TN$ respectively represent the number of false positives and true negatives. Therefore, the bi-objective optimization problem is formally defined as:

\begin{equation}
\begin{aligned}
\max \quad & f_1(\mathbf{x}) = \text{AUPRC} \\
\min \quad & f_2(\mathbf{x}) = \text{FPR}
\end{aligned}
\end{equation}

The NSGA-III framework is then employed to optimize these conflicting objectives and maintain diversity along the Pareto front.

\textit{2) Semantically Aligned Feature Encoding}

Before detailing the individual representation and to lay the groundwork for valid cross-task transfer within the upcoming ENM, a unified representation space is strictly required. We propose a Semantically Aligned Encoding Scheme anchored on the fixed lexicographical order of the 15 nucleotide types ($N_0, \dots, N_{14}$). Unlike relative indexing, the feature search space (indices $0-28$, derived from Eq. 5) is organized by global biological identity. For any target task $\tau \in \mathcal{T}$, the feature pool is structured as follows:

\begin{algorithm}[h]
\caption{Framework of ENM}
\label{alg:neighborhood}
\renewcommand{\algorithmicrequire}{\textbf{Input:}}
\renewcommand{\algorithmicensure}{\textbf{Output:}}
\begin{algorithmic}[1]
\REQUIRE 
    Populations $\{P_{t}\}$ and fitness for all tasks $t \in \mathcal{T}$;
    Neighborhood size $K$; Threshold $\rho$.
\ENSURE External-Neighborhood sets $\mathbf{N} = \{\mathcal{N}_t\}_{t \in \mathcal{T}}$.
\STATE Initialize global elite pool $\mathcal{E}_{ext} \leftarrow \emptyset$;
\FOR{each task $t \in \mathcal{T}$}
    \STATE Select elite individuals $E_{t}$ from $P_{t}$ based on primary objective;
    \STATE $\mathcal{E}_{ext} \leftarrow \mathcal{E}_{ext} \cup E_{t}$;
\ENDFOR
\FOR{each task $t \in \mathcal{T}$}
    \STATE $\mathcal{N}_t \leftarrow \emptyset$;
    \FOR{each individual $p \in P_t$}
        \STATE Initialize similarity list $L_p \leftarrow \emptyset$;
        \FOR{each elite $e \in \mathcal{E}_{ext}$ from tasks $t' \neq t$}
            \STATE Calculate Gray Relation Grade $\gamma(p, e)$;
            \STATE Add $(e, \gamma(p, e))$ to $L_p$;
        \ENDFOR
        \STATE Sort $L_p$ in descending order of $\gamma$;
        \STATE Select top-$K$ elites from $L_p$ to form $\mathcal{N}_p$;
    \ENDFOR
    \STATE Add $\mathcal{N}_t$ to $\mathbf{N}$;
\ENDFOR
\RETURN $\mathbf{N}$;
\end{algorithmic}
\end{algorithm}
\label{subsec:fow}

Indices 0--14 (Target as Primary): $\tau$ serves as the primary task, fused with each nucleotide $k \in \{0-14\}$ in the fixed set acting as auxiliary context. Notably, the index $k$ where $N_k = \tau$ denotes the single-task embedding (Self-Fusion), while others represent specific dual-task embeddings.

Indices 15--28 (Target as Auxiliary): This range implements semantic role reversal, where $\tau$ serves as the auxiliary context supporting the other 14 nucleotides (sequentially ordered, excluding $\tau$).

This alignment enforces strict topological consistency: Index 0 invariably encodes the interaction with the first nucleotide (e.g., ADP), independent of the target task's identity. This invariance ensures that beneficial interaction patterns identified by the evolutionary algorithm—such as leveraging ADP context—are semantically preserved during architectural migration.

\textit{3) Individual Representation and Feature Fusion}

Each individual employs a Feature-Operator-Weight (f-o-w) encoding comprising a subset of feature indices, a sequence of operators, and continuous weights. Let $\mathbf{E}^{(0)}(\mathbf{X})$ denote accumulated representation and $\mathbf{E}^{(j)}(\mathbf{X}) \in \mathcal{S}_{\tau}$ (Eq. 2)  the next selected feature matrix. The recursive fusion step is defined as:
\begin{equation}
\mathbf{E}^{(0)}(\mathbf{X}) \leftarrow w_c \cdot \mathbf{E}^{(0)}(\mathbf{X}) \ \circledast \ w_f \cdot \mathbf{E}^{(j)}(\mathbf{X})
\end{equation}
where $\circledast \in \{\texttt{add}, \texttt{mul}, \texttt{max}, \texttt{min}, \texttt{diff}, \texttt{avg}\}$. Sequential application of this operator over the selected subset yields the final fused matrix $\mathbf{E}^{(fused)}(\mathbf{X})$.

\textit{4) Optimization Process}

The optimization framework (outlined in Algorithm~\ref{alg:main_framework}) commences with random population initialization. In each generation, we first construct an external-neighborhood via GRA to facilitate valid cross-task knowledge transfer (detailed in Section \ref{subsec:external_neighborhood}). Parents are then sampled from either the current task population or this constructed neighborhood, guided by the aligned encoding.

To rigorously explore the search space, we apply a multi-level mutation strategy: Structural Mutation modifies the architecture topology (insertion/deletion/replacement); Operator Mutation updates interaction logic; and Weight Mutation perturbs continuous parameters. Crucially, to prevent the rejection of promising architectures due to suboptimal initialization, we incorporate a Batch Differential Evolution (Batch DE) mechanism. This mechanism employs a probabilistic DE/current-to-best/1 strategy. This lifetime learning step fine-tunes the continuous weights of offspring prior to fitness evaluation, ensuring architectures are assessed based on their potential optimal parameterization. Finally, the population is updated via NSGA-III environmental selection.

\subsection{ENM Method}
\label{subsec:external_neighborhood}

Motivated by the shared physicochemical mechanisms (e.g., hydrogen bonding, electrostatics, and stacking) of structurally related ligands like ATP and GTP~\cite{Zhang2022PanSpecific}, we posit an architectural inductive bias: tasks governing biologically similar binding mechanisms imply similar optimal pathways for feature extraction. Consequently, we design the ENM that uses the Gray Relational Grade (GRG)~\cite{Kandula2023Gray} to explicitly measure strategy similarity. By constructing a reservoir of high-quality architectural motifs from related tasks, this mechanism allows the algorithm to transfer elite ``structural blueprints'' rather than searching from scratch, thereby facilitating collaborative exploration of the combinatorial search space.

Inspired by MTEA/D-DN~\cite{9721409}, our approach employs a unified feature--operator--weight (f-o-w) encoding (see Section~\ref{subsec:fow}) to enable direct similarity evaluation across tasks. We quantify individual similarity via GRG and construct the external-neighborhood for each task by selecting the most similar elite individuals from other tasks.

\textit{1) GRG Calculation}

Given individual vectors $\mathbf{x}=[x_1,\ldots,x_n]$ and $\mathbf{y}=[y_1,\ldots,y_n]$, we first compute the absolute deviation $\Delta_i = |y_i - x_i|$ for each dimension. Let $\Delta_{\min}$ and $\Delta_{\max}$ denote the minimum and maximum deviations across all dimensions. The Gray Relational Coefficient (GRC) for the $i$-th dimension is defined as:
\begin{equation}
\zeta_i = \frac{\Delta_{\min} + \rho \Delta_{\max}}{\Delta_i + \rho \Delta_{\max}}
\end{equation}
where the distinguishing coefficient $\rho$ is set to $0.25$. The GRG is then calculated as the mean GRC:
\begin{equation}
\gamma(\mathbf{y},\mathbf{x}) = \frac{1}{n}\sum_{i=1}^{n}\zeta_i
\end{equation}

\textit{2) Elite Selection and ENM Process}

To mitigate the computational cost of exhaustive GRG calculation, we restrict the candidate pool to the elite subset of each task, defined as the set of top-performing individuals based on the primary objective (i.e., validation AUPRC). For a given individual, its external-neighborhood is formed by selecting the top-$K$ most similar elites from other tasks based on GRG ranking (see Algorithm~\ref{alg:neighborhood}).

\begin{table}[htbp]
  \centering
  \caption{Detailed statistics of the Nuc-1982 dataset distribution.}
  \label{tab:dataset_stats}
  \renewcommand{\arraystretch}{1} %稍微增加行高，让表格不拥挤
  \setlength{\tabcolsep}{8pt}
  
  \begin{tabular}{lcccc}
    \toprule
    \multirow{2}{*}{\textbf{Type}} & \multicolumn{2}{c}{\textbf{Training and Validation Set}} & \multicolumn{2}{c}{\textbf{Test Set}} \\
    \cmidrule(lr){2-3} \cmidrule(lr){4-5}
     & (Pos, Neg)$^a$ & Ratio$^b$ & (Pos, Neg)$^a$ & Ratio$^b$ \\
    \midrule
    ATP & (4688, 183273) & 2.55\% & (1403, 84267) & 1.66\% \\
    ADP & (5145, 210355) & 2.44\% & (1131, 68974) & 1.64\% \\
    AMP & (2505, 100429) & 2.49\% & (240, 12958)  & 1.85\% \\
    GDP & (1523, 56859)  & 2.67\% & (347, 16795)  & 2.06\% \\
    GTP & (1204, 50405)  & 2.38\% & (462, 19723)  & 2.34\% \\
    TMP & (224, 7652)    & 2.93\% & (31, 762)     & 4.07\% \\
    CTP & (484, 20888)   & 2.32\% & (152, 8706)   & 1.74\% \\
    CMP & (315, 8273)    & 3.81\% & (42, 3130)    & 1.34\% \\
    UTP & (444, 19712)   & 2.25\% & (56, 5801)    & 0.96\% \\
    UMP & (103, 2398)    & 4.29\% & (38, 1745)    & 2.18\% \\
    UDP & (974, 37356)   & 2.61\% & (230, 11954)  & 1.92\% \\
    IMP & (209, 6826)    & 3.06\% & (64, 2143)    & 2.99\% \\
    GMP & (178, 7060)    & 2.52\% & (9, 316)      & 2.85\% \\
    CDP & (311, 10922)   & 2.85\% & (58, 2559)    & 2.27\% \\
    TTP & (347, 14685)   & 2.36\% & (104, 7340)   & 1.42\% \\
    \bottomrule
    \multicolumn{5}{l}{\footnotesize$^a$Number of positive (Pos) and negative (Neg) samples.} \\
    \multicolumn{5}{l}{\footnotesize$^b$Ratio of positive samples.} \\
  \end{tabular}
\end{table}

During offspring generation, the external-neighborhood serves as a reservoir of diverse genetic material. By allowing interactions between the current task population and its external-neighborhood during crossover and weight mutation operations—for instance, pairing a parent from the current task with a high-quality neighbor from a related task—this strategy introduces structurally valid yet diverse fusion patterns, thereby enhancing the algorithm's global search capability and stability.

\subsection{Test Process}
\label{subsec:test_process}
Upon conclusion of the evolutionary algorithms, the optimal fusion strategy is identified from the final Pareto front based on validation metrics and instantiated for inference. For a query protein sequence, the inference process commences by generating a candidate feature pool via the pretrained PLMs and the trained multi-granularity encoder. Guided by the decoded optimal strategy, specific feature subsets are selected, standardized, and sequentially fused according to the learned operators and weights. The resulting representation is mapped to logit scores via a pre-optimized coefficient matrix derived from the classification head, followed by a sigmoid activation to yield the final residue-level binding probabilities.

\section{EXPERIMENT DESIGN}

\subsection{Datasets}

To rigorously evaluate model robustness and generalization under realistic scenarios, we employ the widely recognized and comprehensive Nuc-1982 benchmark dataset. This dataset encompasses 15 nucleotide-binding tasks, comprising five high-resource nucleotides (e.g., ATP, GTP) and ten low-resource nucleotides.

\begin{table}[htbp]
    \centering
    \caption{Hyperparameters used in MTGA-MGE}
    \label{tab:hyperparameters}
    \renewcommand{\arraystretch}{1}
    \setlength{\tabcolsep}{9pt}
    \begin{tabular}{ll}
        \toprule
        \textbf{Hyperparameter} & \textbf{Setting} \\ 
        \midrule
        \multicolumn{2}{c}{\textbf{Stage I: Feature Extraction}} \\
        \midrule
        \textit{Model Architecture} & \\
        Input Dimension & 2304 (ProtT5 + ESM2) \\
        CNN Kernel Sizes & \{1, 3, 7\} \\
        CNN Channel Dimensions & 512 $\to$ 256 $\to$ 128 \\
        Transformer Heads & 8 \\
        Transformer Hidden Dim & 128 \\
        Dropout Rate & 0.5 \\
        \textit{Optimization \& Setup} & \\
        Optimizer & AdamW \\
        Learning Rate & $1 \times 10^{-4}$ \\
        Weight Decay & $1 \times 10^{-4}$ \\
        Batch Size & 16 \\
        Maximum Epochs & 30 \\
        Early Stopping Patience & 5 \\
        Validation Set Ratio & 25\% \\
        \textit{Loss Function} & \\
        Type & Focal Loss \\
        Focusing Parameter ($\gamma$) & 1.5 \\
        Class Balancing Weights ($\alpha$) & Negative: 0.15, Positive: 0.85 \\
        \midrule
        \multicolumn{2}{c}{\textbf{Stage II: Evolutionary Algorithms}} \\
        \midrule
        \textit{Evolutionary Strategy} & \\
        Optimizer & NSGA-III \\
        Population Size & 50 \\
        Number of Generations & 40 \\
        Crossover Probability & 0.9 \\
        Mutation Probability & 0.6 \\
        Maximum Feature Length & 25 \\
        Transfer Neighborhood Size ($K$) & 10\% of Population \\
        Gray Relation Coefficient ($\rho$) & 0.25 \\
        \textit{Evaluation Head (Logistic)} & \\
        Maximum Iterations & 300 \\
        Ridge Regularization ($\lambda$) & 0.5 \\
        \bottomrule
    \end{tabular}
\end{table}

Data Partitioning and Preprocessing: We adopted a strict temporal split strategy to simulate predictions on future data. The training and validation sets were derived from proteins released prior to October 26, 2023. To mitigate bias arising from homologous proteins, sequences within each nucleotide category were redundancy-reduced at a 30\% sequence identity threshold. For the three data-heavy tasks (ADP, AMP, ATP), we implemented a 25\% stratified subsampling to balance computational efficiency with representativeness, while retaining all available data for the remaining tasks. Preliminary experiments showed that this subsampling ratio provided sufficient representativeness for convergence without compromising performance stability. Ultimately, 25\% of the processed data was held out as the validation set. The independent test set consists of proteins released on or after October 26, 2023. This time-based partitioning ensures zero data leakage during evaluation, thereby objectively reflecting the model's generalization boundaries.

Distributional Challenge: The benchmark is characterized by extreme Class Imbalance (Table \ref{tab:dataset_stats}). Positive binding residues account for only $\sim$2.77\% and 2.09\% of the training and independent test sets, respectively. This sparse distribution poses a severe challenge for the model to capture faint binding signals while maintaining a low false positive rate.

\begin{table*}[htbp]
\centering
\caption{Performance Comparison with State-of-the-Art Methods on Five High-resource Nucleotide Tasks}
\label{tab:nucleotide_comparison}
\renewcommand{\arraystretch}{0.9}
\setlength{\tabcolsep}{6pt}
\begin{tabular*}{\textwidth}{@{\extracolsep{\fill}} clcccccc}
\toprule
\textbf{Nucleotide} & \textbf{Method} & \textbf{MCC} & \textbf{AUPRC} & \textbf{Sen (\%)} & \textbf{Pre (\%)} & \textbf{Acc (\%)} & \textbf{Spe (\%)} \\
\midrule
\multirow[t]{6}{*}{ATP}
 & SXGBsite   & 0.343 & 0.377 & 52.49 & 24.62 & 96.16 & 96.98 \\
 & EC-RUS     & 0.423 & 0.311 & 48.77 & 38.78 & 97.64 & 98.55 \\
 & BGSVM-NUC  & 0.357 & 0.350 & 45.94 & 30.01 & 97.03 & 97.99 \\
 & GHKNN      & 0.120 & 0.223 & 46.09 & 5.67  & 84.89 & 85.62 \\
 & DeepATPseq      & 0.531 & 0.506 & 49.11 & 58.94  & 98.61 & \textbf{99.43} \\
 & NucGMTL    & 0.570 & 0.552 & 55.38 & 60.10 & 98.66 & 99.38 \\
 & MTGA-MGE  & \textbf{0.667} & \textbf{0.716} & \textbf{69.85} & \textbf{64.86} & \textbf{98.89} & 99.37 \\
\midrule
\multirow[t]{6}{*}{ADP}
 & SXGBsite   & 0.387 & 0.425 & 57.19 & 28.03 & 96.87 & 97.53 \\
 & EC-RUS     & 0.484 & 0.390 & 55.85 & 43.55 & 98.08 & 98.78 \\
 & BGSVM-NUC  & 0.405 & 0.390 & 50.22 & 34.65 & 97.61 & 98.41 \\
 & GHKNN      & 0.162 & 0.314 & 51.30 & 7.37  & 88.55 & 89.17 \\
 & NucGMTL    & 0.604 & 0.599 & \textbf{59.84} & 62.32 & 98.76 & 99.40 \\
 & MTGA-MGE  & \textbf{0.684} & \textbf{0.674} & 50.13 & \textbf{94.19} & \textbf{99.15} & \textbf{99.95} \\
\midrule
\multirow[t]{6}{*}{AMP}
 & SXGBsite   & 0.279 & 0.269 & 50.00 & 17.99 & 94.95 & 95.78 \\
 & EC-RUS     & 0.392 & 0.259 & 37.50 & 43.06 & 97.96 & 99.08 \\
 & BGSVM-NUC  & 0.252 & 0.224 & 29.58 & 24.15 & 97.03 & 98.27 \\
 & GHKNN      & 0.111 & 0.194 & 33.33 & 6.44  & 89.98 & 91.02 \\
 & NucGMTL    & 0.555 & 0.572 & 50.58 & \textbf{62.62} & \textbf{98.55} & \textbf{99.44} \\
 & MTGA-MGE  & \textbf{0.619} & \textbf{0.648} & \textbf{67.50} & 58.06 & 98.52 & 99.10 \\
\midrule
\multirow[t]{6}{*}{GDP}
 & SXGBsite   & 0.488 & 0.466 & 50.15 & 49.85 & 97.57 & 98.74 \\
 & EC-RUS     & 0.506 & 0.499 & 36.17 & 73.01 & 98.13 & 99.67 \\
 & BGSVM-NUC  & 0.494 & 0.447 & 39.82 & 63.59 & 98.00 & 99.43 \\
 & GHKNN      & 0.323 & 0.423 & 50.46 & 23.71 & 94.87 & 95.97 \\
 & NucGMTL    & 0.661 & 0.654 & \textbf{61.56} & 72.24 & 98.74 & 99.51 \\
 & MTGA-MGE  & \textbf{0.695} & \textbf{0.754} & 61.38 & \textbf{79.78} & \textbf{98.90} & \textbf{99.68} \\
\midrule
\multirow[t]{6}{*}{GTP}
 & SXGBsite   & 0.357 & 0.362 & 40.04 & 34.64 & 96.90 & 98.23 \\
 & EC-RUS     & 0.399 & 0.384 & 41.34 & 41.16 & 97.30 & 98.62 \\
 & BGSVM-NUC  & 0.415 & 0.352 & 32.25 & 56.02 & 97.87 & \textbf{99.41} \\
 & GHKNN      & 0.270 & 0.349 & 40.04 & 21.39 & 95.26 & 96.55 \\
 & NucGMTL    & 0.572 & 0.594 & 50.48 & \textbf{66.72} & 98.29 & \textbf{99.41} \\
 & MTGA-MGE  & \textbf{0.649} & \textbf{0.675} & \textbf{65.58} & 65.87 & \textbf{98.43} & 99.20 \\
\bottomrule
\end{tabular*}
\end{table*}

\begin{figure*}[!htbp]
    \centering
    \includegraphics[width=\textwidth]{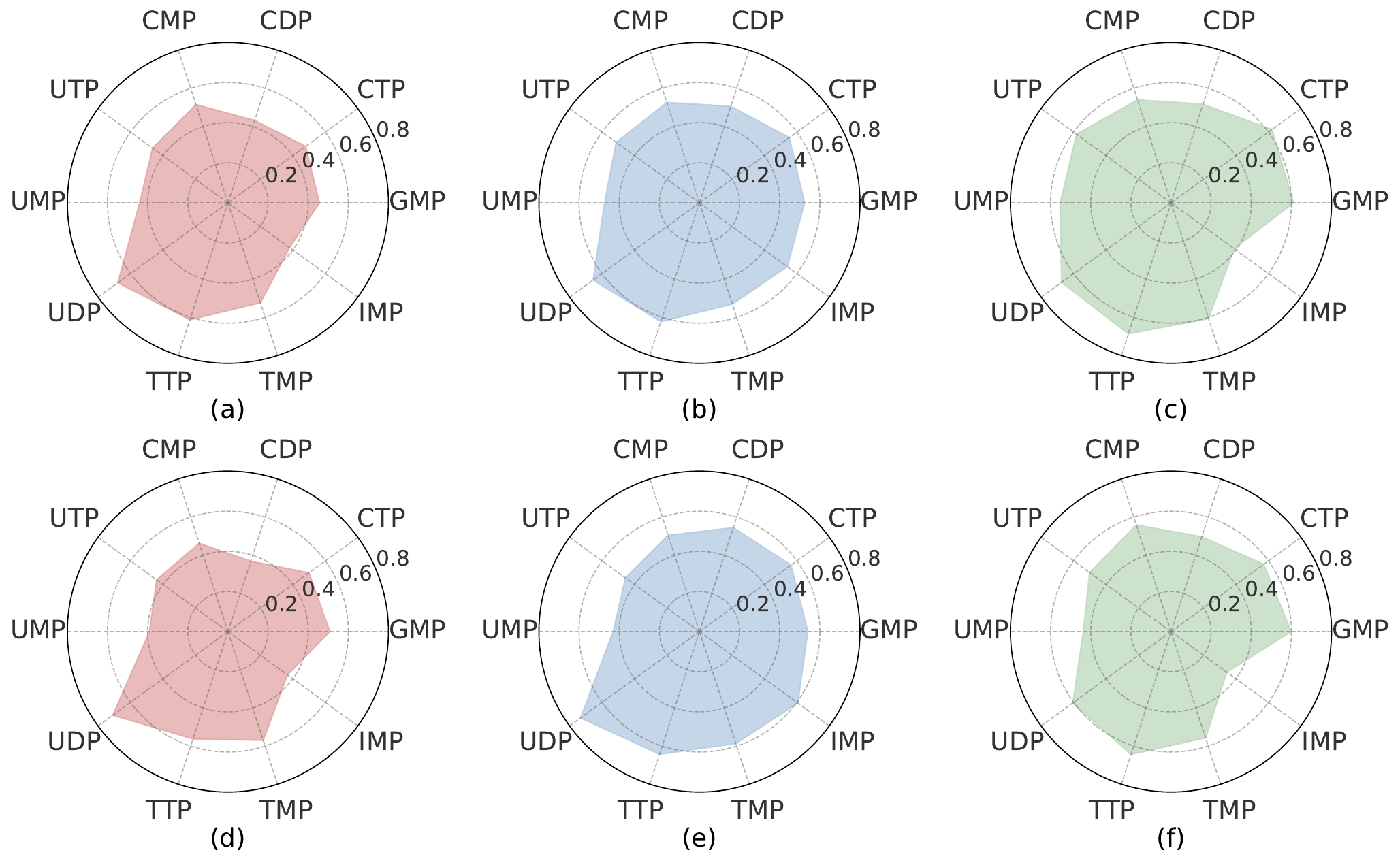}
    \caption{Radar plots comparing the performance of NucMTL (red), NucGMTL (blue), and MTGA-MGE (green) on ten low-resource nucleotide tasks. Panels (a)--(c) display the MCC, while panels (d)--(f) present the AUPRC. In each plot, the axes represent specific nucleotide tasks, with a larger enclosed area indicating superior predictive performance.}
    \label{fig:nucleotide_radar}
\end{figure*}

\begin{figure*}[htbp]
    \centering
    % ================= 第一排 4 张图 =================
    \begin{subfigure}{0.23\linewidth}
        \centering
        \includegraphics[width=\linewidth]{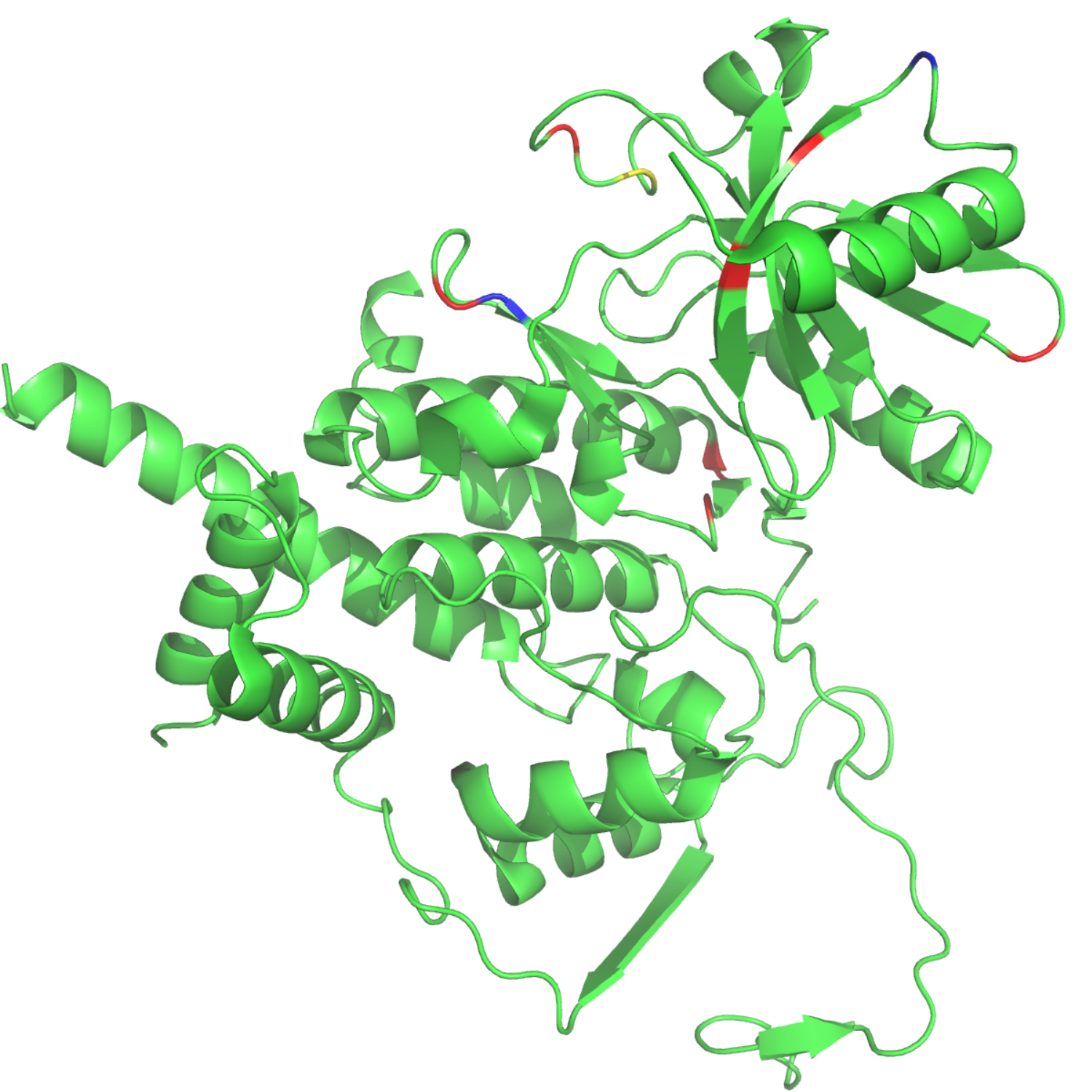}
        \caption{7pt7\_8}
        \label{fig:7pt7_8}
    \end{subfigure}% <-- 注意这里的百分号很重要，它能消除换行带来的多余空格
    \hspace{0.2cm}% <-- 这里控制图片左右的绝对距离，你可以随意把 0.5 改成 0.2 或 1
    \begin{subfigure}{0.23\linewidth}
        \centering
        \includegraphics[width=\linewidth]{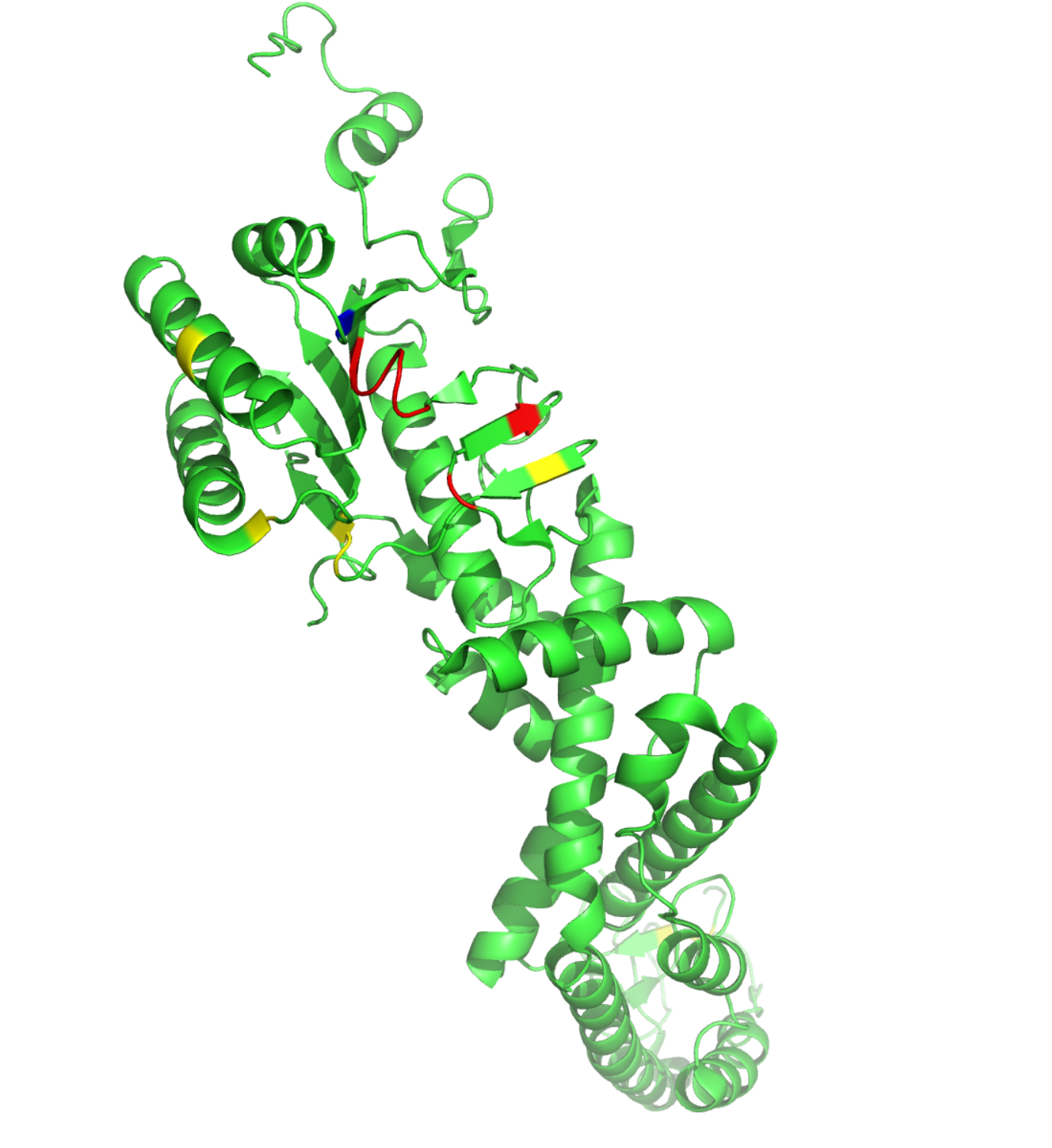}
        \caption{7vuk\_B}
        \label{fig:7vuk_B}
    \end{subfigure}%
    \hspace{0.2cm}%
    \begin{subfigure}{0.23\linewidth}
        \centering
        \includegraphics[width=\linewidth]{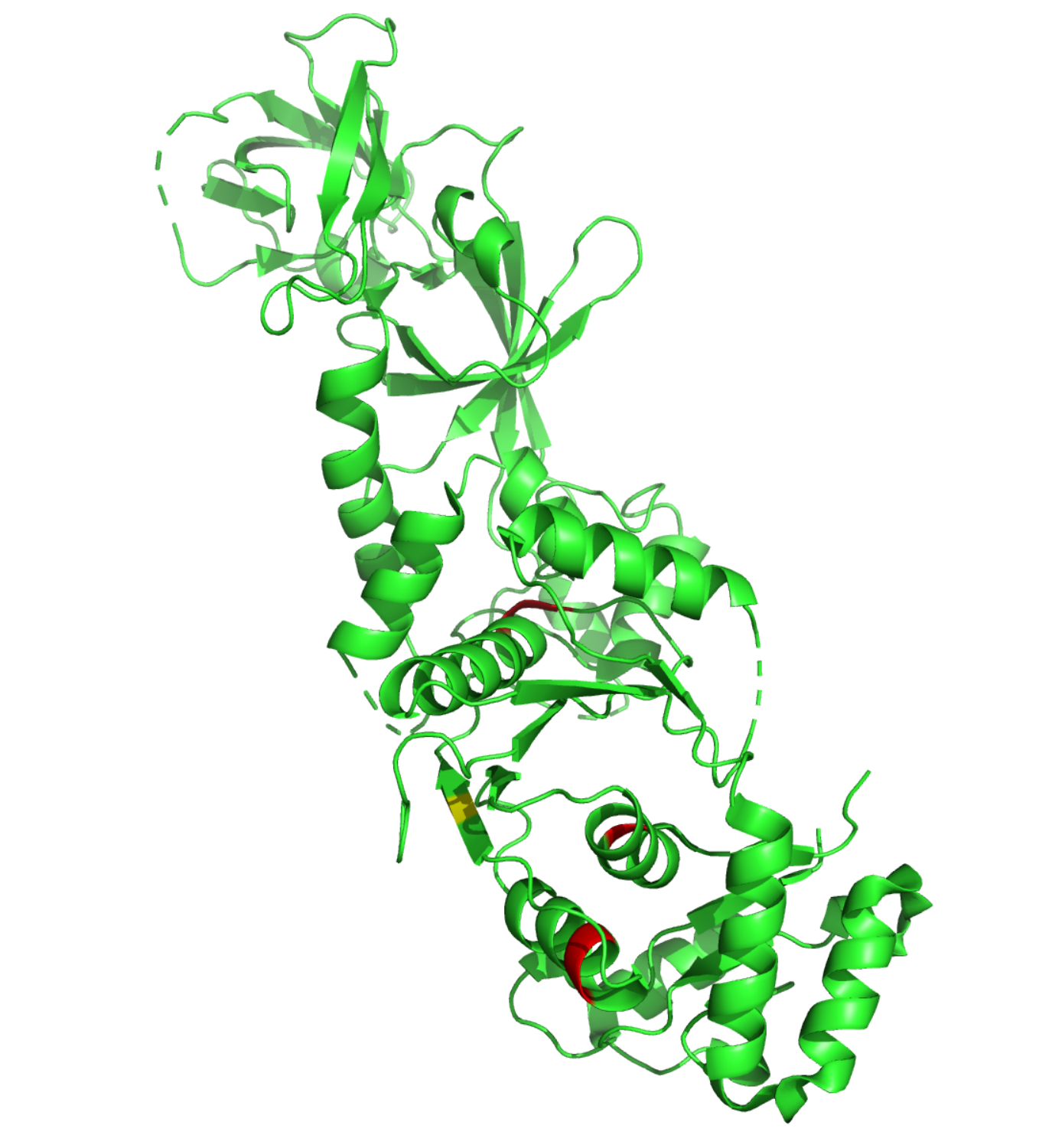}
        \caption{7wak\_A}
        \label{fig:7wak_A}
    \end{subfigure}%
    \hspace{0.2cm}%
    \begin{subfigure}{0.23\linewidth}
        \centering
        \includegraphics[width=\linewidth]{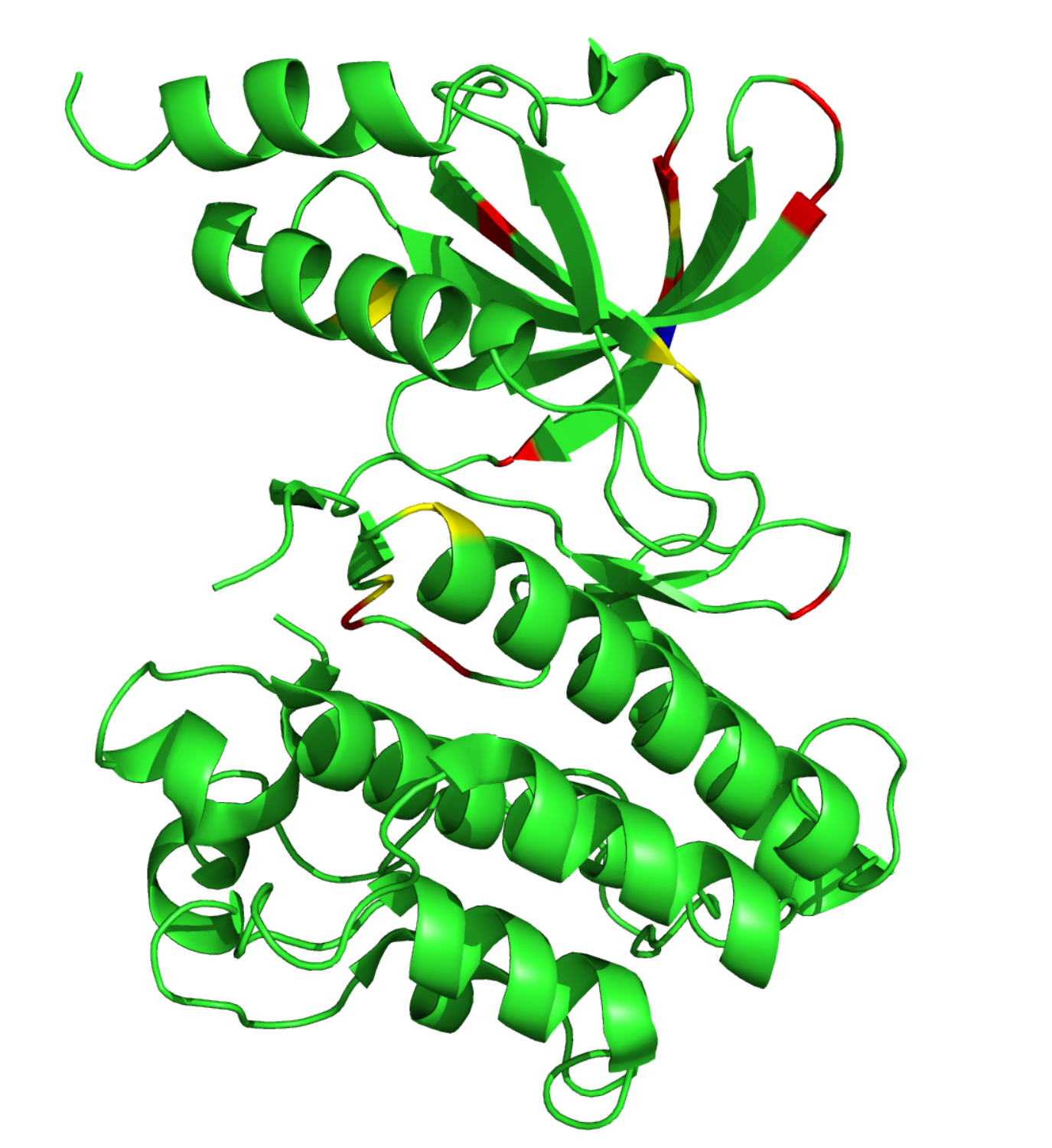}
        \caption{7zvs\_A}
        \label{fig:7zvs_A}
    \end{subfigure}
    
    \vspace{1em} % 上下两排的垂直间距
    
    % ================= 第二排 4 张图 =================
    \begin{subfigure}{0.23\linewidth}
        \centering
        \includegraphics[width=\linewidth]{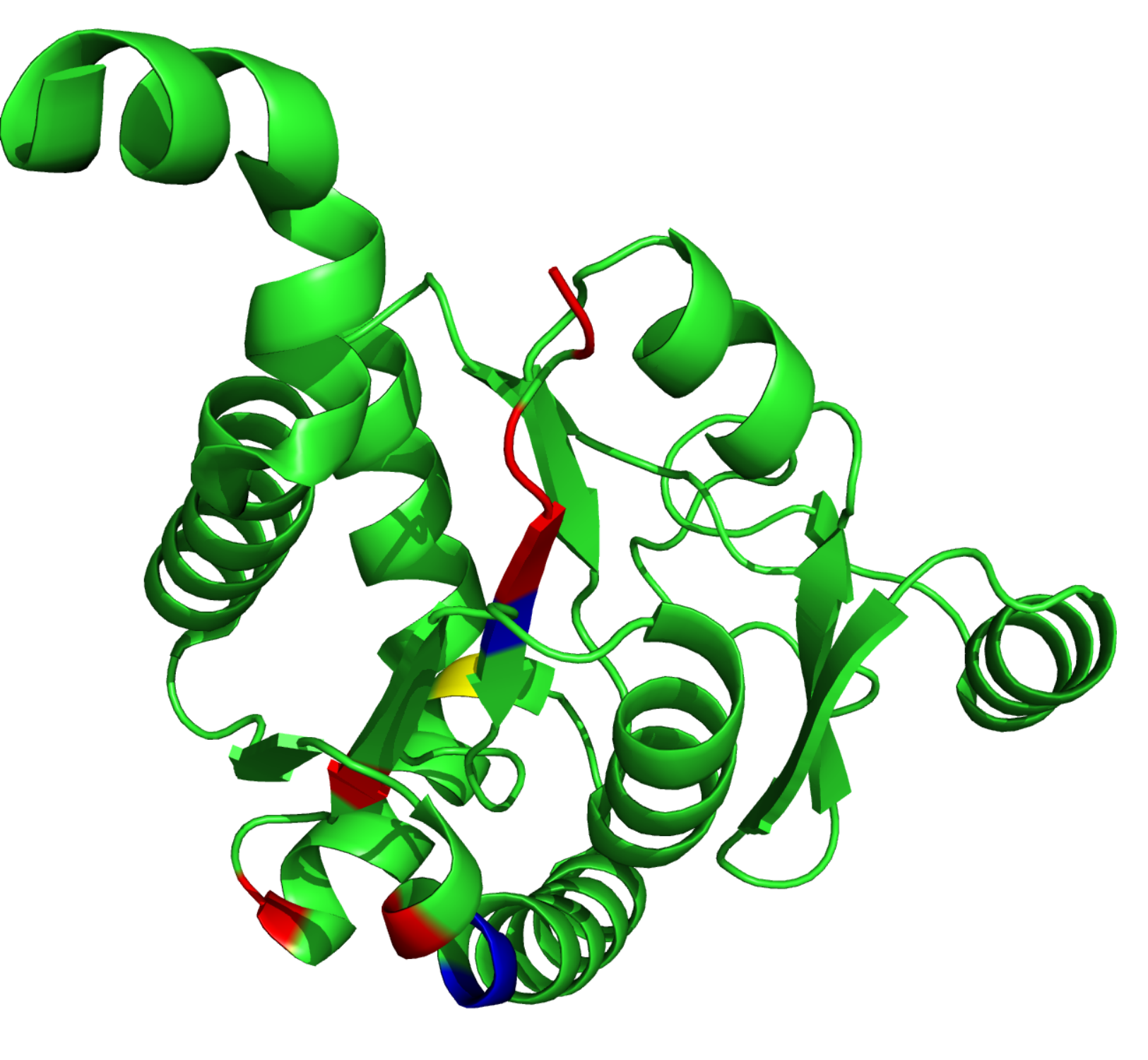}
        \caption{7s5e\_A}
        \label{fig:7s5e_A}
    \end{subfigure}%
    \hspace{0.2cm}%
    \begin{subfigure}{0.23\linewidth}
        \centering
        \includegraphics[width=\linewidth]{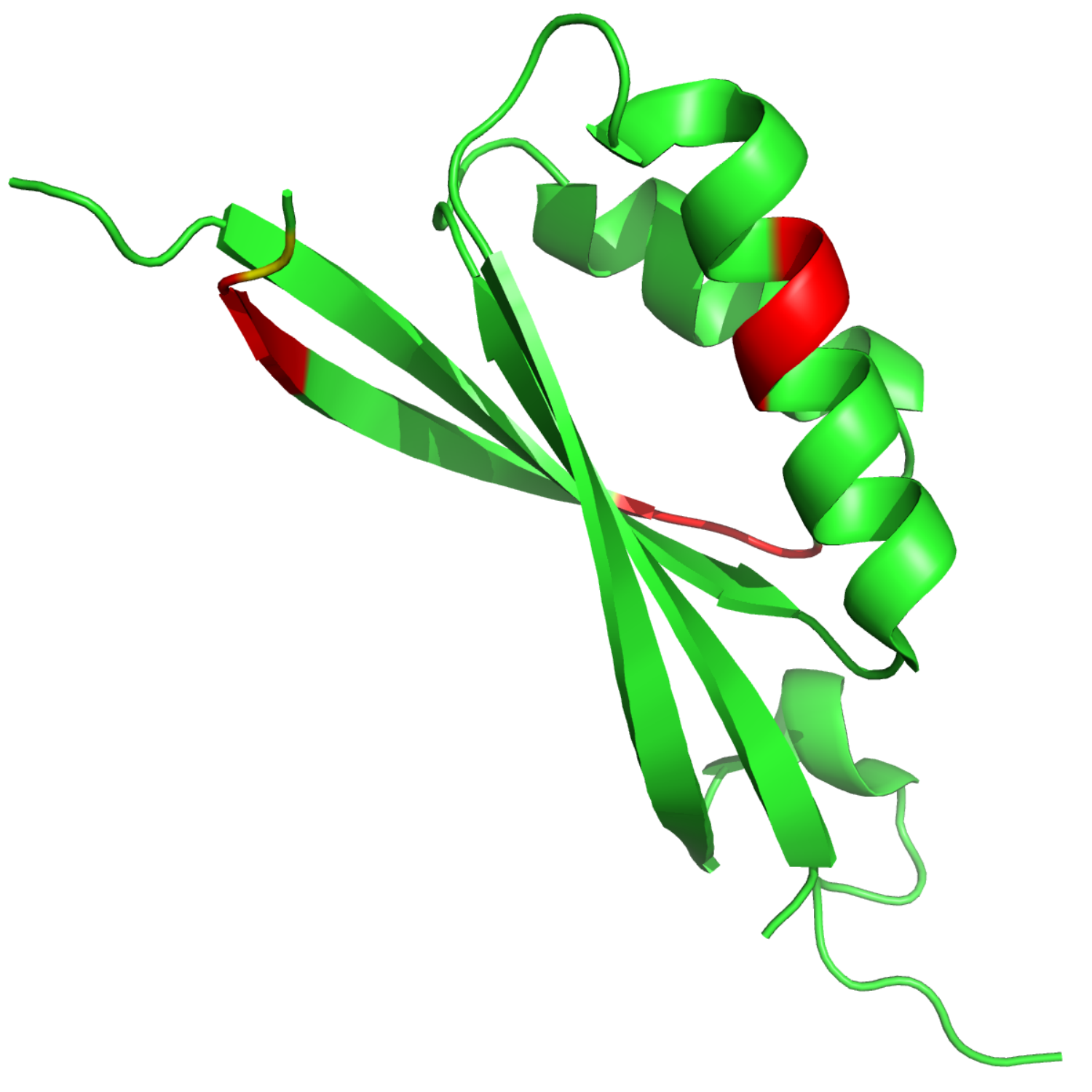}
        \caption{7r30\_B}
        \label{fig:7r30_B}
    \end{subfigure}%
    \hspace{0.2cm}%
    \begin{subfigure}{0.23\linewidth}
        \centering
        \includegraphics[width=\linewidth]{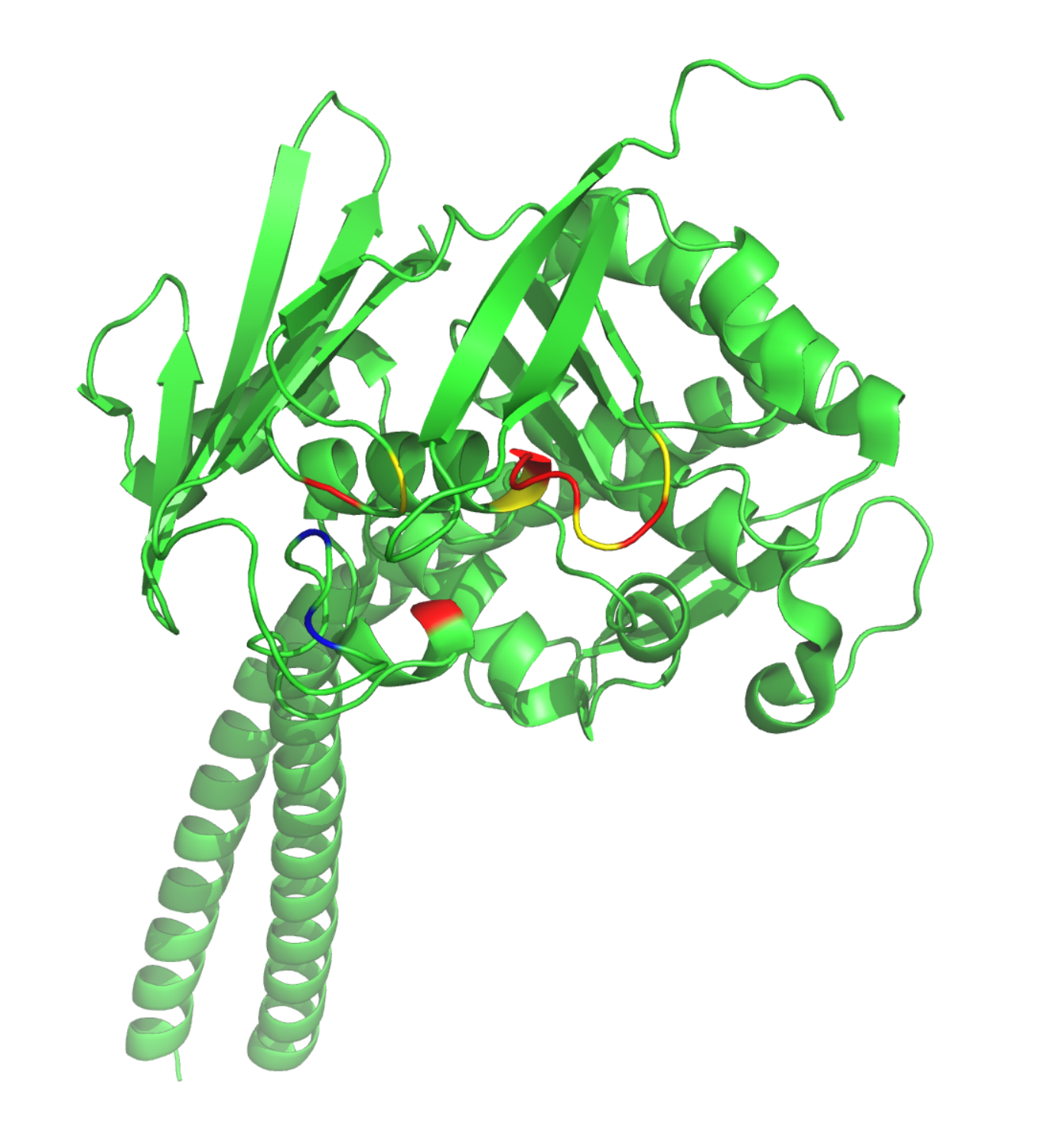}
        \caption{7w1m\_A}
        \label{fig:7w1m_A}
    \end{subfigure}%
    \hspace{0.2cm}%
    \begin{subfigure}{0.23\linewidth}
        \centering
        \includegraphics[width=\linewidth]{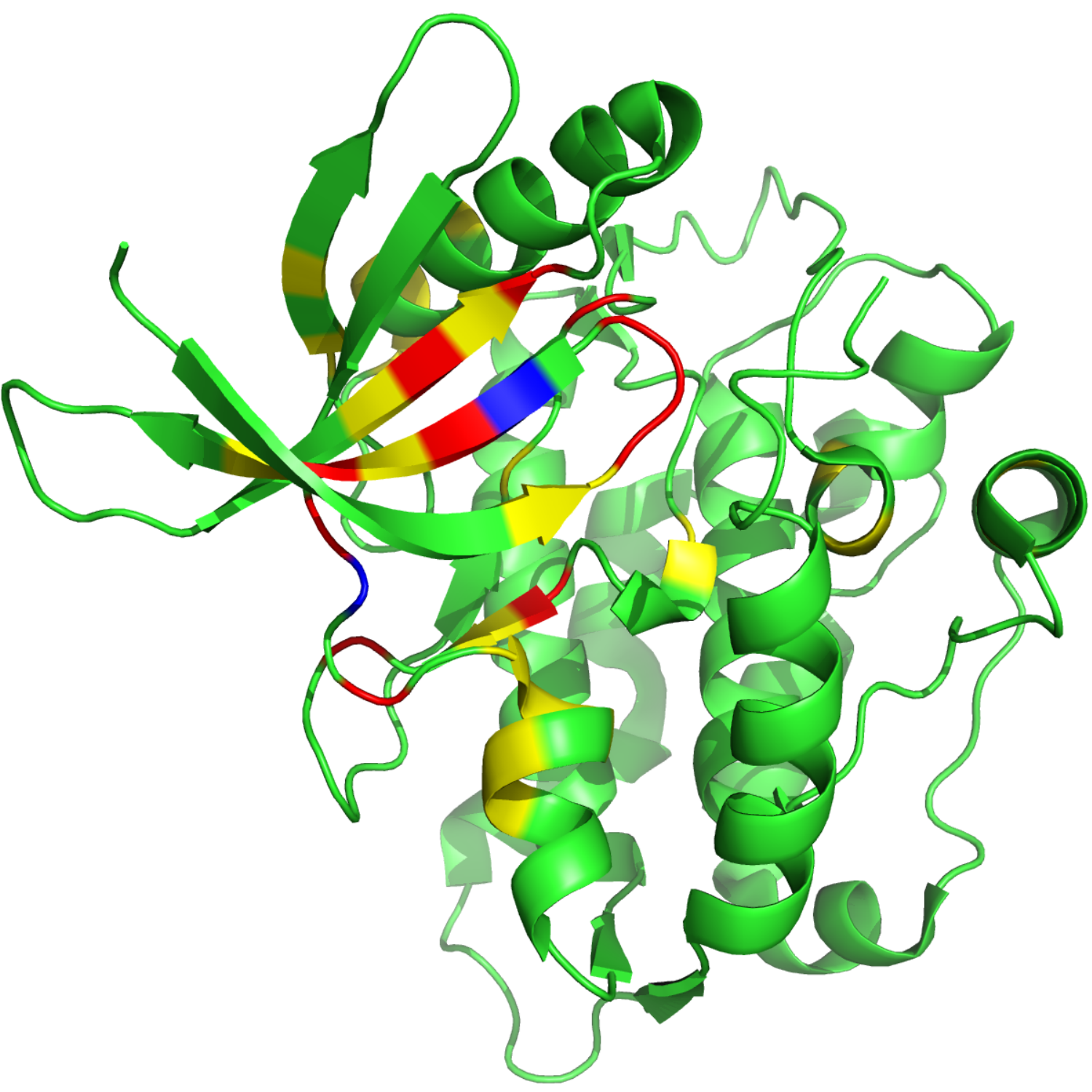}
        \caption{8ak3\_A}
        \label{fig:8ak3_A}
    \end{subfigure}
    
    \caption{Visualization of protein-ADP binding site predictions. The protein structures are shown in gray cartoon representation. Red: True Positive (TP); Blue: False Negative (FN); Yellow: False Positive (FP).}
    \label{fig:comparison}
    \rule{\textwidth}{0.75pt}
\end{figure*}

\subsection{Parameter Settings}
All experiments were conducted using the PyTorch framework on a single NVIDIA GeForce RTX 4090 GPU. In the feature extraction stage, the encoder is optimized using AdamW, employing Focal Loss to address the severe class imbalance. For the evolutionary stage, we utilize the NSGA-III algorithm to balance feature dimensionality and predictive accuracy. Detailed configurations for both model architecture and optimization strategies are comprehensively listed in Table \ref{tab:hyperparameters}.

\subsection{Evaluation}
The model's performance is assessed using six key metrics: AUPRC,
Matthews Correlation Coefficient (MCC), Sensitivity (Sen), Precision (Pre), Specificity (Spe), and Accuracy (Acc).

AUPRC and MCC: AUPRC and MCC are prioritized as the primary metrics, given that the extreme class imbalance (average positive ratio $< 3\%$) renders standard accuracy unreliable for residue-level prediction. AUPRC quantifies the trade-off between precision and recall across decision thresholds:
\begin{equation}
\label{eq:auprc}
\mathrm{AUPRC} = \int_{0}^{1} P(R) \, dR
\end{equation}
\noindent where $P(R)$ denotes precision as a function of recall $R$. Additionally, MCC provides a robust quality measure even for highly unbalanced classes by utilizing all confusion matrix entries:

\begin{equation}
\label{eq:mcc}
\mathrm{MCC} = \frac{TP\cdot TN - FP\cdot FN}{\sqrt{(TP+FP)(TP+FN)(TN+FP)(TN+FN)}}
\end{equation}

\noindent where $TP$, $TN$, $FP$, and $FN$ represent true positives, true negatives, false positives, and false negatives, respectively.

Supplementary Metrics: To ensure comprehensive evaluation, we also report Sensitivity, Precision, Specificity, and Accuracy, calculated as follows:

\begin{equation}
\label{eq:Sen}
\mathrm{Sen} = \frac{TP}{TP+FN}
\end{equation}

\begin{equation}
\label{eq:Pre}
\mathrm{Pre} = \frac{TP}{TP+FP}
\end{equation}

\begin{equation}
\label{eq:Spe}
\mathrm{Spe} = \frac{TN}{TN+FP}
\end{equation}

\begin{equation}
\label{eq:Acc}
\mathrm{Acc} = \frac{TP+TN}{TP+TN+FP+FN}
\end{equation}

\section{Experimental Results}

\subsection{Comparison with State-of-the-Art Methods}

We rigorously benchmarked MTGA-MGE against six representative nucleotide-general models and several nucleotide-specific predictors across 15 tasks in the Nuc-1982 suite. The evaluation covered both high-resource and low-resource nucleotides, with detailed results presented in Table~\ref{tab:nucleotide_comparison} and Fig.~\ref{fig:nucleotide_radar}. Additionally, the spatial distribution of predictions is qualitatively illustrated in Fig.~\ref{fig:comparison}, where the predicted residues (red) accurately delineate the physical binding pockets.

\begin{figure*}[!htbp]
    \centering
    \includegraphics[width=\textwidth, trim=0 17cm 0 0, clip]{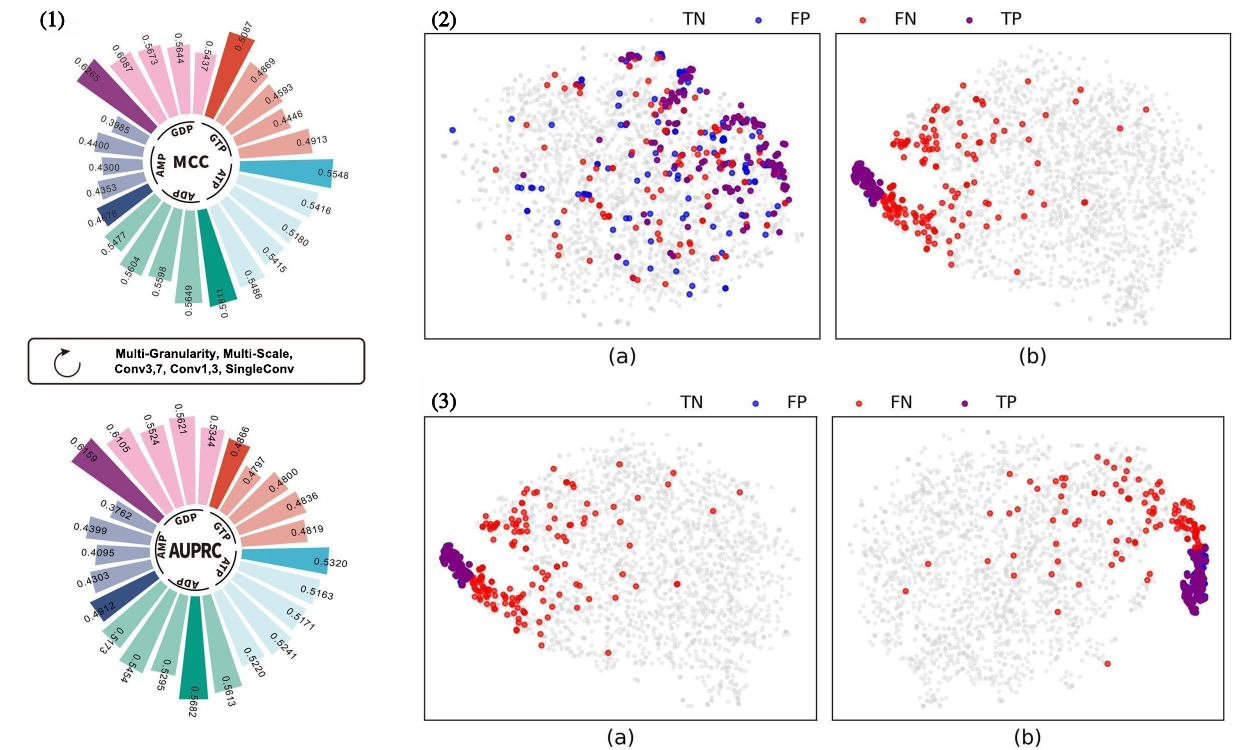}
    \caption{Overall performances of MGE. (1) Performance comparison (MCC and AUPRC) of different convolutional encoder architectures across five high-resource nucleotide tasks. (2) t-SNE visualization comparing the residue representations learned (a) using raw PLM features alone and (b) using MGE. (3) t-SNE visualization comparing residue representations learned under different input configurations: (a) single-task GDP input and (b) joint GDP--ADP input. In all t-SNE plots, data points are colored by prediction outcome: true positive (TP, purple), false negative (FN, red), false positive (FP, blue), and true negative (TN, gray).}
    \rule{\textwidth}{0.75pt}
    \label{fig:MGE}
\end{figure*}

Performance on High-resource Nucleotide Tasks: MTGA-MGE achieved the highest AUPRC and MCC scores across all five high-resource nucleotide tasks, significantly outperforming state-of-the-art competitors. As shown in Table~\ref{tab:nucleotide_comparison}, our model surpassed the nearest baseline by an average margin of 0.099 in AUPRC and 0.070 in MCC. Task-specific analysis reveals that for AMP, the high AUPRC despite lower precision indicates an effective ranking of true binding residues; conversely, for ADP, the combination of high precision and robust sensitivity suggests the establishment of strict and accurate decision boundaries within highly skewed distributions.

Performance on Low-resource Nucleotide Tasks: Despite extreme data scarcity, MTGA-MGE demonstrated competitive adaptability when compared to both the traditional multi-task learning baseline (NucMTL) and the state-of-the-art NucGMTL (Fig.~\ref{fig:nucleotide_radar}). Among the ten low-resource nucleotide tasks, the model secured performance gains on six, avoiding performance collapse under low-resource conditions. Notably, even in the extremely data-scarce UTP and CTP tasks, MTGA-MGE maintained competitive AUPRC benchmarks.

\subsection{Effectiveness of MGE}

To validate the architectural advantage of the MGE, we conducted a controlled ablation study treating the encoder as the sole variable. Under a unified training protocol, we benchmarked our MGE against three representative variants: a single-scale baseline, a serial stacked configuration, and a dual-branch parallel architecture. This setup disentangles the contribution of different receptive field modeling strategies to representation quality.

As illustrated in Fig.~\ref{fig:MGE}(1), quantitative analysis confirms the consistent superiority of the MGE across all five high-resource nucleotide tasks. The single-scale baseline exhibited the poorest performance, verifying that limited receptive fields are inadequate for capturing complex binding cues under extreme class imbalance. While serial and simple dual-branch configurations offer marginal improvements, MGE achieves the highest MCC and AUPRC scores. This demonstrates that the parallel integration of complementary semantics from short-, medium-, and long-range contexts is optimal for robust feature extraction.

\begin{figure*}[!htbp]
    \centering
    \includegraphics[width=\textwidth, trim=0 0 0 0, clip]{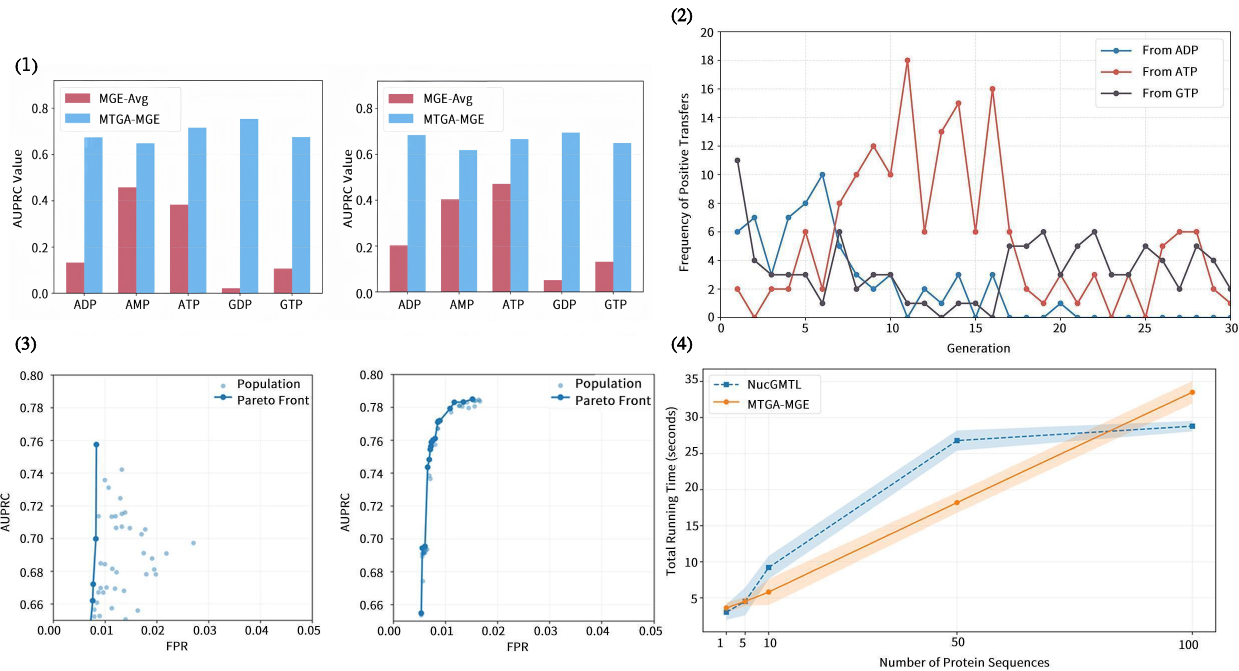}
    \caption{Overall performance of MTGA optimization and MTGA-MGE inference efficiency. (1) Performance comparison between Naive Mean Fusion (MGE-Avg) and GA (MTGA-MGE) on five high-resource nucleotide tasks. (2) Temporal evolution of interaction intensity for the GDP task across evolutionary generations. (3) Evolution of the Pareto front for the GDP task. The left panel shows the initial population with random scattering, and the right panel illustrates the final population converging toward a non-dominated front. (4) Comparison of inference efficiency between NucGMTL and MTGA-MGE on the ADP task across different batch sizes.}
    \rule{\textwidth}{0.75pt}
    \label{fig:MTGA}
\end{figure*}

Visualization Analysis: To intuitively elucidate the encoder's role, we employed t-SNE to visualize the latent representations (Fig.~\ref{fig:MGE}(2)). Compared to raw PLM embeddings, the refined features exhibit markedly enhanced class separability, characterized by distinct clustering of positive and negative samples. This qualitatively corroborates the encoder's efficacy in filtering high-dimensional background noise. Notably, the error distribution reveals 
a specific ``halo effect'': false positives are not randomly scattered but are spatially clustered in the immediate vicinity of true binding sites. This phenomenon reflects an intrinsic trade-off in convolutional modeling: while multi-scale aggregation amplifies weak binding signals (thereby boosting recall), it inevitably blurs local boundaries via spatial smoothing. These ``near-misses'' indicate that the model successfully localizes binding pockets despite imprecise boundary delineation, suggesting future avenues for fine-grained boundary refinement.

\subsection{Effectiveness of Dual-task Context Injection}

To validate the benefits of the proposed multi-task context injection, we compared feature representations derived from single-task GDP versus joint GDP--ADP inputs. As visualized in Fig.~\ref{fig:MGE}(3), the joint-input configuration qualitatively exhibits markedly clearer class separation, an observation rigorously corroborated by quantitative metrics computed on the original high-dimensional embeddings. The average inter-class distance increased from $10.86$ to $13.81$, while the Davies--Bouldin (DB) score—where a lower value indicates superior discriminability based on the ratio of intra-class compactness to inter-class separation—improved from $0.948$ to $0.721$. These results confirm that joint-task training effectively leverages cross-task semantic correlations to enhance feature discriminability.

This performance gain can be attributed to the underlying biological homology. Since homologous ligands (e.g., GDP and ADP) typically target pockets with similar physicochemical properties, we hypothesize that the auxiliary task provides a critical ``structural reference'' during training. For residues exhibiting ambiguous signals in the target task, the shared auxiliary context likely facilitates the alignment of their features toward regions with well-defined binding patterns. This process appears to effectively mitigate feature divergence in single-task scenarios, transforming loose, task-specific distributions into a more coherent and collaborative representation.

\subsection{Impact of GA}

To assess the contribution of GA, we benchmarked the full MTGA-MGE framework against a non-evolutionary baseline, MGE-Avg, where the adaptive genetic algorithm is replaced by a static arithmetic mean strategy across all 29 candidate feature sources. As visualized in Fig.~\ref{fig:MTGA}(1), replacing evolutionary selection with naive aggregation induces severe performance degradation across all high-resource nucleotide tasks. This failure is particularly catastrophic for GDP and GTP; notably, MGE-Avg achieves a negligible MCC of 0.051 on GDP (compared to 0.695 for MTGA-MGE), essentially losing all discriminative capability.

\begin{table}[htbp]
\centering
\caption{Performance Comparison between With and Without the ENM on Five High-resource Nucleotides}
\label{tab:external_neighborhood}
\renewcommand{\arraystretch}{1.1} % 增加行高
\small % 小号字体
\setlength{\tabcolsep}{0pt} % 让 extracolsep 自动填充
\begin{tabular*}{\columnwidth}{@{\extracolsep{\fill}} @{\hspace{5pt}} c l c c @{\hspace{5pt}}}
\toprule
\textbf{Nucleotide} & \textbf{Optimization Strategy} & \textbf{MCC} & \textbf{AUPRC} \\
\midrule
\multirow{2}{*}{ADP} & w/o-neighborhood & \textbf{0.596} & 0.611 \\
                     & neighborhood    & 0.593 & \textbf{0.630} \\
\midrule
\multirow{2}{*}{AMP} & w/o-neighborhood & 0.567 & 0.530 \\
                     & neighborhood    & \textbf{0.618} & \textbf{0.600} \\
\midrule
\multirow{2}{*}{ATP} & w/o-neighborhood & 0.653 & 0.699 \\
                     & neighborhood    & \textbf{0.677} & \textbf{0.715} \\
\midrule
\multirow{2}{*}{GDP} & w/o-neighborhood & 0.653 & 0.670 \\
                     & neighborhood    & \textbf{0.700} & \textbf{0.742} \\
\midrule
\multirow{2}{*}{GTP} & w/o-neighborhood & 0.586 & 0.603 \\
                     & neighborhood    & \textbf{0.637} & \textbf{0.665} \\
\bottomrule
\end{tabular*}
\end{table}

This collapse highlights the inherent risks of static multi-task aggregation. By indiscriminately averaging the target task with 28 auxiliary contexts, the baseline treats all information sources as equally relevant. Consequently, the distinctive feature patterns required for identifying specific binding residues are likely compromised by irrelevant or conflicting information from biologically distinct tasks. In contrast, MTGA-MGE’s superiority stems from its capacity for autonomous selective fusion. By dynamically retaining only compatible feature subsets, the evolutionary framework effectively filters out interference while amplifying target signals. These results confirm that the adaptive selection provided by the evolutionary module, rather than mere information accumulation, is indispensable for resolving task heterogeneity and ensuring robust performance.

\subsection{Impact of ENM}

To evaluate the contribution of ENM, we conducted a controlled ablation study on five high-resource nucleotide tasks. We contrast the baseline configuration lacking the external mechanism (denoted as w/o-neighborhood) against the enhanced strategy incorporating cross-task neighborhoods (denoted as neighborhood). In the w/o-neighborhood setting, each task evolves independently based solely on its own population; conversely, the neighborhood setting permits offspring to sample high-quality parents from the external-neighborhood, facilitating selective cross-task genetic variation. All other experimental configurations were kept identical to ensure rigorous comparability. As evidenced in Table~\ref{tab:external_neighborhood}, neighborhood consistently outperforms w/o-neighborhood in AUPRC and MCC metrics across four of the five tasks, with only marginal fluctuations observed for ADP. These results suggest that by exploiting complementary search trajectories from biologically related tasks, the neighborhood strategy effectively bolsters the model's capacity to detect rare binding residues under conditions of extreme class imbalance.

To elucidate the underlying dynamics of this gain, we monitored the temporal evolution of interaction intensity for the GDP task (Fig.~\ref{fig:MTGA}(2)). The process exhibits distinct phasic characteristics: initial generations show high-frequency interactions with external tasks (ADP, ATP, GTP), indicating that the ENM actively leverages diversity to escape local optima during the early exploration phase. As evolution proceeds, reliance on cross-task transfer diminishes in favor of task-specific refinement. This transition confirms that the mechanism facilitates a natural shift from global collaborative exploration to local exploitation, autonomously regulating the dependency on information transfer as the population converges.
 
\begin{table}[htbp]
\centering
\caption{Performance Comparison of Different Multi-Objective Optimization Strategies on Four Representative High-resource Nucleotide Tasks}
\label{tab:optimizer_comparison}
\renewcommand{\arraystretch}{1} % 增加行高
\small % 小号字体
\setlength{\tabcolsep}{0pt} % 让 extracolsep 自动填充
\begin{tabular*}{\columnwidth}{@{\extracolsep{\fill}} @{\hspace{7pt}} c l c c @{\hspace{7pt}}}
\toprule
\textbf{Nucleotide} & \textbf{Method} & \textbf{MCC} & \textbf{AUPRC} \\
\midrule
\multirow{3}{*}{ADP} & SPEA2    & 0.610 & 0.603 \\
                     & NSGA-II  & 0.586 & 0.544 \\
                     & NSGA-III & \textbf{0.684} & \textbf{0.674} \\
\midrule
\multirow{3}{*}{AMP} & SPEA2    & 0.681 & 0.646 \\
                     & NSGA-II  & 0.659 & 0.643 \\
                     & NSGA-III & \textbf{0.684} & \textbf{0.648} \\
\midrule
\multirow{3}{*}{GDP} & SPEA2    & 0.645 & 0.585 \\
                     & NSGA-II  & 0.694 & 0.729 \\
                     & NSGA-III & \textbf{0.700} & \textbf{0.742} \\
\midrule
\multirow{3}{*}{GTP} & SPEA2    & 0.581 & 0.574 \\
                     & NSGA-II  & 0.574 & 0.565 \\
                     & NSGA-III & \textbf{0.592} & \textbf{0.601} \\
\bottomrule
\end{tabular*}
\end{table}

\subsection{Comparison of Different Multi-Objective Optimizers}

To justify the selection of NSGA-III as the core optimization engine, we conducted a controlled comparison against two established baselines: NSGA-II, which relies on crowding distance for diversity maintenance, and SPEA2, which utilizes a strength-based fitness and external archive mechanism. Quantitative results across four representative high-resource nucleotide tasks (ADP, AMP, GDP, GTP), summarized in Table~\ref{tab:optimizer_comparison}, demonstrate that NSGA-III consistently yields the highest MCC and AUPRC scores. This superiority is likely attributed to the unique reference-point-based mechanism of NSGA-III. Unlike the crowding distance estimation of NSGA-II or the nearest-neighbor density estimation of SPEA2, the reference-point strategy provides a more structured and uniform exploration of the solution space, making it particularly robust for navigating the complex discrete combinatorial landscape of feature fusion.
 
Fig.~\ref{fig:MTGA}(3) tracks the evolution of the solution space for the GDP task. The transition from an initially scattered distribution to a condensed, continuous frontier empirically validates the algorithm's convergence capability. This progression from a diffuse cloud to a well-defined frontier demonstrates effective selection pressure, successfully driving the population toward non-dominated solutions that represent optimal trade-offs between sensitivity (AUPRC) and specificity (FPR) for downstream model selection.

\subsection{Inference Efficiency Analysis}

To evaluate the practical utility of MTGA-MGE, we benchmarked its inference latency against the state-of-the-art NucGMTL model. The ADP-binding task was selected as the representative benchmark because its number of fused features lies in the median range among all evaluated nucleotides, providing a balanced perspective on computational complexity and model performance. Although MTGA-MGE involves an evolutionary search during the training phase, the final inference is optimized by pruning the computational graph to execute only the task-specific optimal paths. The empirical results, as illustrated in Fig. \ref{fig:MTGA}(4), demonstrate that MTGA-MGE maintains high efficiency across different batch sizes. Notably, for medium-scale inputs (10–50 sequences), MTGA-MGE even outperforms NucGMTL, completing 50 sequences in 18.36 seconds compared to 26.80 seconds. These results verify that our proposed multi-granularity encoding and adaptive architecture do not impose a significant computational burden during deployment, ensuring its viability for high-throughput protein-nucleotide binding site prediction.

\section{Discussion}

Based on comprehensive evaluations across the Nuc-1982 benchmark and various cross-ligand scenarios, we conclude that the MTGA-MGE framework offers distinct advantages in three key aspects: (1) Superior Predictive Robustness and Generalization. By transitioning from manual feature fusion to an adaptive optimization paradigm, the model effectively breaks the limitations inherent in fixed architectures. This flexibility ensures that the learned representations remain robust when faced with the complex structural characteristics of low-resource nucleotides, thereby effectively enhancing the model's overall generalization capability. (2) Enhanced Representation Purity and Signal Sensitivity. This study achieves a significant leap in representational clarity by precisely distilling core binding signals from complex backgrounds. By refining the granularity of feature extraction, the model filters out high-dimensional noise that often obscures critical residues. This refinement heightens the model's sensitivity to the subtle spatial arrangements of binding sites, enabling more precise localization of interaction anchors within intricate protein structures. (3) Homology-Driven Biological Synergy. The framework effectively bridges the gap between isolated tasks by explicitly exploiting physicochemical homologies. By transforming disjoint, task-specific distributions into structurally aligned representations, the model allows the shared biological context to calibrate latent features. This ``co-evolutionary'' design facilitates structural information complementarity across tasks, thereby stabilizing the predictive performance of various ligands and maximizing the overall robustness of the framework.

\section{Conclusion}

In this study, we addressed the challenges of task heterogeneity in protein-nucleotide binding site prediction by proposing MTGA-MGE, a unified framework that transitions from rigid, manual feature fusion to an autonomous, adaptive architecture search. Extensive evaluations demonstrate that MTGA-MGE establishes a new state-of-the-art on high-resource nucleotide tasks while maintaining competitiveness on low-resource ones. These results validate that modeling feature fusion as a discrete combinatorial optimization problem is a powerful approach to navigating the semantic space of high-dimensional multi-source representations.

Despite these strengths, we acknowledge certain limitations that warrant further investigation. First, the evolutionary process incurs high computational overhead during training due to the iterative nature of the search. However, this cost is strictly confined to the offline stage; once the architecture is optimized, online inference remains exceptionally fast and viable for high-throughput screening. Second, the model remains sensitive to extreme data scarcity. Effective transfer requires minimal distributional overlap, and in regimes of extreme sparsity, stochastic noise may outweigh transfer benefits. Future work will address these constraints by integrating structural priors or few-shot learning techniques to elevate performance in low-resource scenarios, and we plan to explore surrogate-assisted optimization to accelerate the search process for broader ligand-binding domains.
% =======================
% ===== 参考文献 ========
% =======================
\bibliographystyle{IEEEtran}
\bibliography{references}
% {\small
% \input{output.bbl}}
\end{document}